\documentclass[10pt,journal,compsoc]{IEEEtran}
\usepackage{comment}

\usepackage{ifpdf}

\ifCLASSOPTIONcompsoc
  \usepackage[nocompress]{cite}
\else
  \usepackage{cite}
\fi
\ifCLASSINFOpdf
   \usepackage[pdftex]{graphicx}
\else
   \usepackage[dvips]{graphicx}
\fi
\usepackage{amsmath}
\usepackage{algorithmic}

\usepackage{array}

\ifCLASSOPTIONcompsoc
  \usepackage[caption=false,font=footnotesize,labelfont=sf,textfont=sf]{subfig}
\else
  \usepackage[caption=false,font=footnotesize]{subfig}
\fi
\usepackage{url}

\newcommand{\mat}[1]{\mathbf{#1}}

\usepackage{pgfplots}

\begin{document}
%
\title{Head2Head++: Deep Facial Attributes Re-Targeting}
%
%
%
%

\author{Michail~Christos~Doukas$_{\ast}^{1,4}$, Mohammad~Rami~Koujan$_{\ast}^{2,4}$, Viktoriia Sharmanska$^1$, Anastasios~Roussos$^{2,3,4}$, and~Stefanos~Zafeiriou$^{1,4}$ 
\IEEEcompsocitemizethanks{
\IEEEcompsocthanksitem $_{\ast}$ denotes equal contribution
\IEEEcompsocthanksitem $^{1}$ Department of Computing, Imperial College London, UK
\IEEEcompsocthanksitem $^{2}$ College of Engineering, Mathematics and Physical Sciences, University of Exeter, UK
\IEEEcompsocthanksitem $^{3}$ Institute of Computer Science (ICS), Foundation for Research and Technology - Hellas (FORTH), GR
\IEEEcompsocthanksitem $^{4}$ FaceSoft.io, London, UK
}

\thanks{Manuscript received June 15, 2020; revised October 3, 2020}}

%
%

\markboth{Journal of T-BIOM,~Vol.~X, No.~X, June~2020}%
{Shell \MakeLowercase{\textit{et al.}}: Bare Demo of IEEEtran.cls for Biometrics Council Journals}
%



\IEEEtitleabstractindextext{%
\begin{abstract}
Facial video re-targeting is a challenging problem aiming to modify the facial attributes of a target subject in a seamless manner by a driving monocular sequence. We leverage the 3D geometry of faces and Generative Adversarial Networks (GANs) to design a novel deep learning architecture for the  task of facial and head reenactment. Our method is different to purely 3D model-based approaches, or recent image-based methods that use Deep Convolutional Neural Networks (DCNNs) to generate individual frames. We manage to capture the complex non-rigid facial motion from the driving monocular performances and synthesise temporally consistent videos, with the aid of a sequential Generator and an ad-hoc Dynamics Discriminator network. We conduct a comprehensive set of quantitative and qualitative tests and demonstrate experimentally that our proposed method can successfully transfer facial expressions, head pose and eye gaze from a source video to a target subject, in a photo-realistic and faithful fashion, better than other state-of-the-art methods. Most importantly, our system performs end-to-end reenactment in nearly real-time speed (18 fps).
\end{abstract}

\begin{IEEEkeywords}
face reenactment, full head reenactment, neural rendering, video renderer, 3DMM, 3D reconstruction, gaze tracking, temporal discriminator, facial flow.
\end{IEEEkeywords}}

\maketitle

\IEEEdisplaynontitleabstractindextext

%
\IEEEpeerreviewmaketitle

\IEEEraisesectionheading{\section{Introduction}\label{sec:introduction}}

%
%
%
%
\IEEEPARstart{I}{mage} and video synthesis receives a rapidly increasing amount of attention in Computer Vision and Deep Learning research, as "synthetic" data appear more and more realistic, as well as especially promising for real-world applications. Video editing, film dubbing, social media content creation, teleconference and virtual assistance are some indicative examples. However, generating artificial human faces indistinguishable from real ones is a very challenging task, particularly when it comes to video data. Adding the extra dimension of time, might give rise to the so-called problem of temporal incoherence. As the uncanny valley effect suggests, people are extremely perceptible in unnatural facial and head movements, thus even small discontinuities can expose a synthetic video.

Over the past years, various methods that target human faces have emerged, taking advantage of different modalities, such as audio \cite{Suwajanakorn2017, X2Face} and video \cite{face2face, deepvideoportraits, koujan_reenactnet} to drive synthesis and dictate the movements of the generated subject. Facial reenactment is a widely-studied approach \cite{thies2015realtime, face2face, DeferredNeuralRendering}, which aims to transfer the facial expressions from a source to a target subject, by conditioning the generative process on the driving video of the source. In most cases \cite{face2face, DeferredNeuralRendering}, this is done by modifying the deformations solely within the internal facial region of the target identity and placing the manipulated face back to the original target frames, using an image interpolation method. Given that face reenactment systems offer no control over the target's head pose and eye gaze, there might be cases where the person's expressions do not match with the overall head movement and therefore seem unnatural. Even so, face reenactment can be very useful for specific applications, such as video dubbing \cite{Garrido2015}, which aims to alter the mouth motion of the target actor to match the audio track spoken by the dubber.

A more holistic approach on reenactment involves generating all pixels within frames, including the upper body, hair and background of the target identity. In contrast to video manipulation techniques, such as face reenactment, full head reenactment methods \cite{deepvideoportraits, fewshot} aim to transfer the entire head motion from a source identity to a target one, along with the eye blinking and gaze, providing complete control over the target subject. Most head reenactment approaches fall into two categories: a) Warping-based methods \cite{Liu2001, Garrido2014, X2Face}, which are mainly learning-based and do not involve computing priors such as facial landmarks or 3D face models, b) object-specific methods \cite{face2face, deepvideoportraits, fewshot, vid2vid}, which assume knowledge (e.g. 3D reconstruction, 68 landmarks) of the source and target faces. On the one hand, warping-based methods usually suffer from distortions in the face and background \cite{X2Face}. On the other hand, object-specific methods relying on facial keypoints are affected from the so-called identity preservation problem. As keypoints encapsulate identity attributes of the source (e.g, head geometry), the more the identity of the source diverges from that of the target, the more distorted the head shape of the generated subject might appear. On the contrary, 3D morphable models (3DMMs) \cite{koujan2018combining, booth20183d} have proven to be a reliable means of decoupling expression and identity from each other. \textit{Deep Video Portraits (DVP)} \cite{deepvideoportraits} is a full head reenactment system that capitalises on 3DMMs and utilises a neural network to translate 3D face reconstructions to realistic frames. Nonetheless, \textit{DVP}, as a purely image-based model, does not take into account temporal dependencies between frames. 

Our recently proposed \textit{Head2Head} \cite{Koujan2020head2head} model overcomes the aforementioned limitations, as it combines the benefits of conditioning synthesis on 3D facial shapes with the advantages of a sequential, video-based, neural renderer. In this paper, we extend the work of \textit{Head2Head} in the following directions:
\begin{itemize}
\item We prevail over the limitations of the 3D reconstruction stage of \textit{Head2Head} by designing a novel \textbf{DenseFaceReg} network for the robust and fast estimation of semantic facial images based on 3D facial geometry. Our semantic facial images, referred to as \textbf{Normalised Mean Face Coordinates (NMFC)} images, capture pose, expression and identity information of the subject from the 3D facial mesh and are used to condition video synthesis.
\item We propose a simple yet fast method for detecting the eye gaze, based on 68 facial landmarks. When combined with 3D Facial Recovery and Video Rendering Network, our system can operate in nearly real-time speeds.
\item We conduct an extensive set of experiments, including user, automated and ablation studies. Our results reveal the significance of video-based modeling.
\item We make our code and dataset publicly available\footnote{\url{https://github.com/michaildoukas/head2head}}.
\end{itemize}

\section{Related Work}


\subsection{3D Face Reconstruction}
\label{subsec: 3Dreconst}

Recovering the 3D geometry of human faces from monocular images is a challenging problem that has attracted much attention due to its major role in many applications, ranging from facial reenactment, performance capture and tracking \cite{Koujan_2020_CVPR}, facial expression recognition \cite{Koujan2020FER, Giannakakis2020}, etc. Owing to their pose and illumination invariance, 3D facial data constitute an indispensable geometrical description of faces for various facial image processing systems. The Computer Vision field is rich in approaches targeting this problem under different assumptions and constraints. Some of these attempts \cite{barron2014shape, smith2008facial}, named \textit{Shape from Shading}, approximate the image formation process and make simplified assumptions about the lighting and illumination models leading to the formation of the image, while others, known as \textit{Structure from Motion (SfM)}, benefit from the geometric constraints in multiple images of the same object to solve the problem \cite{garg2013dense, grasshof2017projective}. One common approach for addressing this task are the \textit{3D Morphable Models (3DMMs)} of the human face. 3DMMs are linear statistical models that have been used substantially since the pioneering work of Blanz and Vetter \cite{blanz1999morphable}, with many extensions \cite{koujan2018combining, booth20183d, faggian20083d}. With the rise of deep neural networks, nonlinear face models have been proposed to recover the 3D geometry and appearance of human faces from images or videos using \textit{deep Convolutional Neural Networks (CNNs)} \cite{tran2019towards, tewari2018self, tran2018nonlinear, tewari2019fml}. For a very recent and comprehensive review of the state-of-the-art methods on monocular 3D face reconstruction and the open challenges of 3DMMs, we refer the readers to \cite{zollhofer2018state, egger20193d}.

In this work, we employ 3DMMs and a 3D shape regression network for: 1) estimating the 3D geometry of faces appearing in videos, and 2) utilising the estimated 3D geometrical information for driving our Deep Video Rendering Neural Network. 

\begin{figure*}[t!]
    \centering
    \includegraphics[scale=1.1]{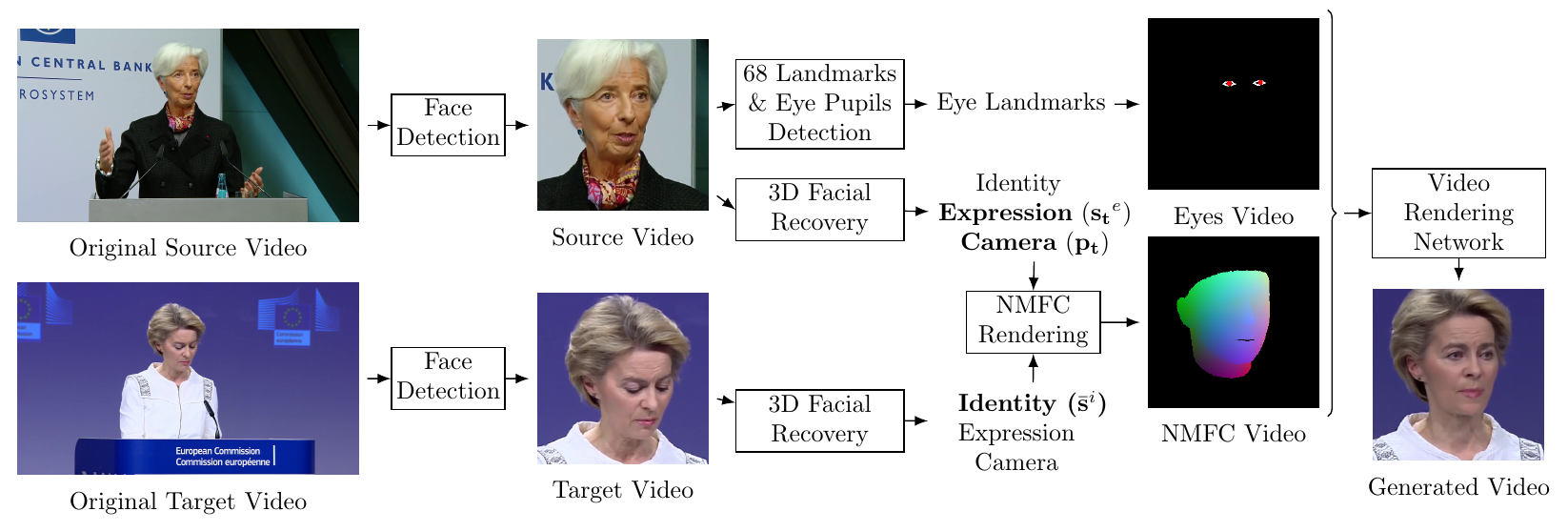}
    \caption{Our Head2Head++ pipeline for head reenactment. First, the facial region of interest is extracted from both the source and target original videos. Next, 3D Facial Recovery is performed. The average identity parameters ($\bar{\mat{s}}^i$) of the target, along with the expression ($\mat{s}_t^e$) and camera ($\mat{p}_t$) parameters of the source for each time-step $t$, are passed to the NMFC renderer, which produces the NMFC video. At the same time, the eye landmarks of the source are used to detect the pupils and sketch the eyes video, which represents eyes movements. Finally, the combined NMFC and eyes video is given as conditional input to the Video Rendering Network, which computes the generated video (reenactment result). The Video Rendering Network is a person specific model, as it is trained using footage of the target person. During training, the source video coincides with the target video, enabling us to perform self-reenactment and get access to ground truth data.}
    \label{fig:fig1}
\end{figure*}

\subsection{Facial Synthesis and Re-targeting}
\label{subsec: animation}


Various deep architectures have been proposed for the image and video synthesis tasks with the aid of Recurrent Neural Networks (RNNs), Variation Auto-encoders (VAE) \cite{kulkarni2015deep}, Gaussian mixture VAE \cite{wang2018every}, Hierarchical VAE \cite{goyal2017nonparametric}, Generative Adversarial Networks (GANs) \cite{ganGoodfellow} and VAE-GAN \cite{Andres}. 

Traditional methods of reenactment, transfer the facial expressions either with 2D warping techniques \cite{Liu2001, Garrido2014}, or by utilizing 3D face models \cite{ thies2015realtime, face2face, DeferredNeuralRendering}. These methods do not provide complete control over the generated video, as they manipulate only the interior of the face. Recently, there has been a substantial effort in the direction of both expression and pose transfer \cite{headon, elor2017bringingPortraits, X2Face, fewshot, deepvideoportraits}. One of the first approaches, \textit{X2Face} \cite{X2Face}, designs an embedding and a set of auto-encoders. \textit{X2face} follows a warping-based approach that causes deformations in the generated heads and inconsistent upper-body motions. Wang et al. \cite{vid2vid} propose \textit{vid2vid}, a GAN-based spatio-temporal approach for the video-to-video synthesis, relying on a Temporal Discriminator for improving the temporal quality of the synthesised videos.
 ~In a follow-up work, Wang et al. \cite{wang2019fewshotvid2vid} extend their approach with an attention mechanism, making it trainable in a few-shot manner and leading to better generalisation performance. 
Both \cite{vid2vid} and \cite{wang2019fewshotvid2vid} can perform reenactment with face sketches drawn from landmarks, leading to a target identity preservation problem. Moreover,. the synthesised mouth regions do not look very realistic.  

Zakharov et al. \cite{fewshot} propose a few-shot, image-based adversarial learning approach, as their network learns to generate unseen target identities even from a single image. Nonetheless, their neural network relies on landmarks, causing an identity distortion of the target. 
Siarohin et al.~\cite{siarohin2019animating} animate objects via a deep motion transfer framework. Given a single image and a driving sequence, their method estimates a dense motion field appearing in the sequence and transfers it to the target image while preserving its appearance. When used for facial reenactment, the faces synthesised by this approach suffer from head distortions and non-naturalistic mouth and teeth areas. To tackle this  issue, the authors extended their work \cite{Siarohin_2019_NeurIPS} to account for complex motions with a first-order motion model and an occlusion-aware Generator. Although this extension  considerably improved the results on various tasks, synthesised faces still exhibited visual artifacts appearing as expression-dependent continuous scale changes. This can be attributed to the estimated 2D dense motion field that does not fully describe the actual intricate 3D facial motion.   

As far as we are aware, \textit{DVP} \cite{deepvideoportraits} is the only learning-based head reenactment system, prior to our work, that uses 3D facial information to condition video synthesis. Their image-based model requires a long video footage of the target person, while training a new model for each target takes many hours. Moreover, generated mouths look unnatural, since the 3D reconstruction method they adopt does not encode the inner-mouth region or the teeth.

Unlike other studies, our approach employs efficiently the 3D geometry and conditions frame generation on a compact and meaningful representation in the image space derived from 3D facial reconstruction. When combined with our novel video-based neural rendering stage, the result is a faithful and photo-realistic full head video reenactment, indistinguishable from real videos. Additionally, we focus specifically on the mouth area and improve its visual quality by designing a dedicated discriminator.


\section{Methodology}

Our \textit{Head2Head++} framework proposes a solution for the largely ill-posed full head reenactment problem. Our method is capable of transferring the time-varying head attributes (pose, expression, eye gaze) and mainly consists of two subsequent modules: a) \textbf{3D Facial Recovery} (Sec.~ \ref{subsec: 3Dtracking}), and b) a \textbf{Video Rendering Network} (Sec. \ref{subsec:videorenderer}). Our video rendering network capitalises on a carefully designed GAN-based framework. Please refer to Fig. \ref{fig:fig1} for an overview of our \textit{Head2Head++} pipeline.

\subsection{3D Facial Recovery}
\label{subsec: 3Dtracking}
We aim to generate a reliable estimation of the facial 3D geometry, capturing the temporal dynamics, while separating the identity and expression contributions of the pictured subject in each frame. We utilise this separability to effectively disentangle the human head characteristics in a transferable and photo-realistic way between different videos. Towards that aim, we benefit from the prior knowledge in our problem space and harness the power of 3DMMs \cite{blanz1999morphable} for reconstructing the faces that appear in the input monocular sequences. Given a sequence of $T$ frames $\mathcal{F}_{1:T}= \{f_t \mid {t=1,\dots,T}\}$, the 3D reconstruction stage produces two sets of parameters: 1) shape parameters $\mathcal{S}=\{\mat{s}_t \mid \mat{s}_t \in {\rm I\!R}^{n_i+n_e}, t=1,...,T\}$, and 2) camera parameters $\mathcal{P}=\{\mat{p}_t \mid \mat{p}_t \in {\rm I\!R}^{6}, t=1,...,T\}$, depicting rotation, translation and orthographic scale. 

\setlength\parindent{0pt} 
\setlength{\parskip}{2pt}

\textbf{Shape representation.} Using 3DMMs, a 3D facial shape $\mathbf{x}_t=[x_1, y_1, z_1,..., x_N, y_N, z_N]^{\top} \in {\rm I\!R}^{3N}$ can be written mathematically as:
\begin{equation}
    \label{eq:3DMM}
\mathbf{x}_t = \mathbf{x}(\mat{s}_t^i,\mat{s}_t^e)=\bar{\mathbf{x}}+\mathbf{U}_{id} \mat{s}_t^i+ \mathbf{U}_{exp} \mat{s}_t^e
\end{equation}
where $\mathbf{\bar{x}}\in {\rm I\!R}^{3N}$ is the mean shape of the morphable model, given by $\mathbf{\bar{x}}=\mathbf{\bar{x}}_{id}+\mathbf{\bar{x}}_{exp}$,  with $\mathbf{\bar{x}}_{id}$ and $\mathbf{\bar{x}}_{exp}$ standing for the mean identity and expression of the model, respectively. $\mathbf{U}_{id} \in {\rm I\!R}^{3N\times n_i}$ is the identity orthonormal basis with $n_i$ principal components ($n_i\ll3N$), $\mathbf{U}_{exp} \in {\rm I\!R}^{3N\times n_e}$ is the expression orthonormal basis with the $n_e$ principal components ($n_e\ll3N$) and $\mat{s}_t^i \in {\rm I\!R}^{n_i}$, $\mat{s}_t^e \in {\rm I\!R}^{n_e}$ are the identity and expression parameters of the morphable model. We designate the joint identity and expression parameters at time-step $t$ by $\mat{s}_t=[\mat{s}_t^{i^{\top}}, \mat{s}_t^{e^{\top}]^{\top}}$.  In the adopted model \eqref{eq:3DMM}, the 3D facial shape $\mathbf{x}$ is a function of both identity and expression coefficients ($\mathbf{x}(\mat{s}_t^i, \mat{s}_t^e)$), where expression variations are effectively represented as offsets from a given identity shape.  

\textbf{Video-based 3D reconstruction.} The video fitting approach followed in \textit{Head2Head} \cite{Koujan2020head2head} to estimate the 3D facial geometry is based on a set of sparse landmarks extracted from the entire input sequence. This method has three main drawbacks: 1) the fidelity of the 3D reconstruction relies heavily on the accuracy of extracted landmarks which are also sparse (68 in total), 2) it might require a large number of frames with enough reconstruction cues (various rotations) to produce good accuracy, 3) it makes a quite strong assumption in the initialisation stage about the rigidity of the face to estimate the camera parameters. To overcome these limitations, we propose to perform the 3D facial reconstruction in this work by training a deep CNN, we call \textbf{DenseFaceReg}, the purpose thereof is to produce a dense 3D facial mesh from a single RGB frame. We use a thousand annotated videos from the \textbf{Face3DVid} dataset of Koujan et al. \cite{Koujan_2020_CVPR} to train this network in a supervised manner. The adopted loss function during training is:

\begin{equation}
    \label{eq:3DFaceMehsReg}
    \begin{split}
        \mathcal{L}(\Phi)= \sum\limits_{i=1}^{N}|| \mathbf{v}_{i}^{GT}-\mathbf{v}_{i}||^2.
    \end{split}
\end{equation}
Equation (\ref{eq:3DFaceMehsReg}) penalises the deviation of each vertex from the corresponding ground-truth vertex ($\mathbf{v}_i=[x, y, z]^{\top}$). We use the camera parameters provided with the \textbf{Face3DVid} dataset to project the 3D ground-truth mesh, so that our \textbf{DenseFaceReg} produces 3D vertices (dense landmarks) directly in the image space. 
We use the dense 3D vertices ($\sim$5K) estimated by our trained \textbf{DenseFaceReg} on each video frame to: 1) estimate the camera parameters, 2) generate the 3DMM identity and expression coefficients by projecting the dense shape onto the 3DMM bases. For all our experiments in this work, we use the same 3DMMs utilised in \cite{Koujan2020head2head}.

The analysis-by-synthesis approach, which is used by many state-of-the-art approaches \cite{deepvideoportraits, face2face}, estimates a lot of parameters (e.g. illumination, reflectance, shape, etc) and solves a highly ill-posed problem for fitting 3DMMs to images. On the contrary, our facial reconstruction stage is a fast CNN-based approach (6ms test runtime) trained on a large number of in-the-wild videos. The facial representation extracted with our method is informative enough to synthesise photo-realistic and temporally smooth videos, eliminating the need for more elaborate and slower 3D facial reconstruction techniques. 

\subsection{Facial Semantic Representation - NMFC Rendering}
Our video rendering network receives as input a facial semantic representation of the target subject, with the head pose and facial expression guided by the source frames. This representation disentangles identity from expression, allowing us to train our video rendering network on a specific target person. During test, we are able to transfer the expression and pose of any source, with different head characteristics, to the target. Given the recovered identity and expression parameters (facial shape) $\mat{s}_t=[\mat{s}_t^{i^{\top}}, \mat{s}_t^{e^{\top}]^{\top}}$ and camera parameters $\mat{p}_t$, at frame $t$, we rasterize the 3D shape, producing a visibility mask ($\mathbf{M} \in {\rm I\!R}^{W\times H}$) in the image space. Each pixel of $\mathbf{M}$, stores the index of the corresponding visible triangle on the 3D face seen from this pixel. Thereafter, we store the normalised x-y-z coordinates of the centre of this triangle in the $\mathbf{NMFC} \in {\rm I\!R}^{W\times H \times3}$ image, which is the facial semantic representation that is utilised as conditional input to the video rendering network. Equation (\ref{eq: NMFC}) details this process.
\begin{equation}
    \label{eq: NMFC}
    \mathbf{NMFC}_t= \mathcal{E}(\mathcal{R}(\mat{x}_t(\mat{s}_t^i, \mat{s}_t^e), \mat{p}_t), \bar{\mat{x}}),
\end{equation}
where $\mathcal{R}$ is the rasterizer, $\mathcal{E}$ is the encoding function and $\bar{\mat{x}}$ is the normalised version of the 3DMM mean face (see (\ref{eq:3DMM})), so that the x-y-z coordinates of this face belong in $[0, 1]$. The NMFC image generation relies not only on the 3D reconstructed face and camera parameters of the current frame but also on an always-fixed normalised mean face ($\bar{\mathbf{x}}$) of the employed 3DMM. The (x, y, z) coordinates of the normalised mean face are used as constant colors to texture the 3D face of the current input frame $f_{t}$. The NMFC image of a frame $f_{t}$ is generated then by rendering the corresponding textured 3D mesh. Since we texture any reconstructed 3D mesh with always a fixed set of distinctive colors, the same semantic point, say the tip of the nose, in any NMFC image (regardless of the subject and the target video) will always have the same color. This is why we refer to NMFCs as semantic images. The main advantage of using NMFCs over a UV 2D parameterisation is that it proved experimentally to be easier for the video renderer to learn the mapping to the output RGB images, since both the NMFCs and the output are in the same space. Additionally, this representation is more compact and easy-to-interpret by our video rendering network, since it associates well with the corresponding RGB frame to be generated, pixel by pixel, and, subsequently, leads to a realistic and novel video synthesis. Note that during test time, the expression coefficients $\mat{s}_t^e$ and camera parameters $\mat{p}_t$ are estimated from the source video at frame $t$, while the identity coefficients $\mat{s}^i$ are the average identity parameters estimated using all frames of the target video (see Fig. \ref{fig:fig1}).

\subsection{Eye Pupils Detection}

We choose a real-time operating method for the extraction of the eye movements from the source frames, which does not require fitting an eye model \cite{Wood2017}, but is based on 68 facial landmarks \cite{dlib09}. Given the subset of landmark points $\mat L^{eye} \in {\rm I\!R}^{6 \times 2}$ that correspond to the left or right eye, we estimate one more landmark $\mat l^{pupil} \in {\rm I\!R}^{2}$, which corresponds to the eye pupil. The eye pupil is obtained as the centre of mass within the set of pixels $\mat \Omega$ that are bounded by the polygon formed with the eye landmarks $\mat L^{eye}$ by computing a weighted sum, using the inverse intensity of pixels in $\mat \Omega$ as weights. Following  \cite{Saragih2011}, we are based on the assumption that eye pupils are the "darker" areas within the eye region in terms of pixel intensity. Mathematically, the eye centre of mass is given as

\begin{equation}
    \mat l^{pupil} = \dfrac{\sum_{\mat p \in \mat \Omega} I(\mat p) \mat p}{\sum_{\mat p \in \mat \Omega} I(\mat p)},
\end{equation}

where $I(\mat p)$ is the inverse intensity of pixel $\mat p$ in the source frame. In this way, $\mat l^{pupil}$ corresponds to the center of "darker" pixels within the eye region.

Once the eyes and pupils landmarks are estimated, we create the eyes video $\mat E_{1:T}$, which is a sequence of $T$ RGB frames of size $256 \times 256 \times 3$. An example of these frames is shown in Fig. 1. We connect the eye landmarks with white edges to create an outline of each eye. Then, two red circles are drawn on the same plane, using as centres the eye pupil coordinates of each eye, in order to indicate eye gaze.

\subsection{Deep Video Rendering Neural Network}

\label{subsec:videorenderer}

Our carefully-designed video rendering network receives as conditional input two sequences of images, namely: the NMFC video $\mathbf{NMFC}_{1:T}$ and the corresponding eye video $\mathbf{E}_{1:T}$, collectively as $\mat{X}_{1:T} \equiv \{ \mat X_t = (\mathbf{NMFC}_t, \mathbf{E}_t) \}_{t=1,\dots,T}$, with $\mat X_t \in {\rm I\!R}^{H \times W \times 6}$. With this input, the video rendering network yields a highly realistic and temporally coherent output video $\tilde{\mat Y}_{1:T}$ picturing the target actor mimicking exactly the same expressions, head pose and eyes reflected in $\mat{X}_{1:T}$. While training, we follow a self reenactment setting where the source and target videos are the same footage. In this way, the reenacted video in the output should be a replication of the RGB training target video $\mat{Y}_{1:T}$, which serves as ground truth. Our video rendering network is trained within a GAN-based structure with: 1) a Generator $G$, which is the video rendering network itself, 2) an Image Discriminator $D_I$, 3) a multi-scale Dynamics Discriminator $D_D$ enforcing the temporal coherence and realism of the produced facial performance, and 4) a dedicated Mouth Discriminator $D_M$, which further improves the visual quality of the mouth area.

\textbf{Generator $G$.} To successfully synthesise smooth and convincing temporal facial performances, we condition the synthesis of the $t$-th frame $\tilde{\mat Y}_t$ not only on the conditional input $\mat X_t$, but also on the previous inputs $\mat X_{t-1}$ and $\mat X_{t-2}$, as well as the previously generated frames $\tilde{\mat Y}_{t-1}$ and $\tilde{\mat Y}_{t-2}$, thus:
\begin{equation}
    \tilde{\mat Y}_{t} = G(\mat{X}_{t-2:t}, \tilde{\mat{Y}}_{t-2:t-1}).
\end{equation}  
The Generator is applied sequentially, producing the output frames one after the other, until the entire output sequence has been created. Similar to \textit{Head2Head} \cite{Koujan2020head2head}, the Generator consists of two identical encoders, operating in parallel, as well as a decoder. The first encoder receives the concatenated NMFC and eye images $\mat{X}_{t-2:t}$, while the second is given the two previously generated frames $\tilde{\mat{Y}}_{t-2:t-1}$. The two extracted feature maps are first added and then passed through the decoder, which brings the output $\tilde{\mat Y}_{t}$ in a normalised [-1,+1] range, using a $\tanh$ activation function.

\textbf{Image Discriminator $D_I$ and Mouth Discriminator $D_M$.} Both of these networks aim at telling real and synthesised frames apart, and are used only during training. At time-step $t$, the Image Discriminator $D_I$ processes the real pair $(\mat X_{t}, \mat Y_{t})$ and the fake one $(\mat X_{t}, \tilde{\mat Y}_{t})$.
Concurrently, the corresponding mouth regions $(\mat X_{t}^m, \mat Y_{t}^m)$ and $(\mat X_{t}^m, \tilde{\mat Y}_{t}^m)$ are cropped and sent to the Mouth Discriminator $D_M$. 

\textbf{Dynamics Discriminator $D_D$.} With the aim of learning complex facial dynamics in mind, we further equip our GAN framework with a Dynamics Discriminator $D_D$. During training, $D_D$ receives a set of three consecutive real frames $\mat{Y}_{t:t+2}$ or fake  frames $\tilde{\mat Y}_{t:t+2}$. They are passed to $D_D$ after being associated with the optical flow $\mat{W}_{1:T-1}$, computed from the real frames (training video $\mat{Y}_{1:T}$ of target subject). Following this, the Generator is encouraged to create fake frames showing the same flow (dynamics) as the corresponding real ones. The Dynamics Discriminator learns to differentiate between the real $(\mat{W}_{t:t+1}, \mat{Y}_{t:t+2})$ and fake $(\mat{W}_{t:t+1}, \tilde{\mat Y}_{t:t+2})$ pairs. In practice, we employ a multiple-scale Dynamics Discriminator, which operates in three different temporal scales. The first scale receives frame sequences in the original frame rate. Then, the two extra scales are formed by sub-sampling the frames by a factor of two for each scale.

\textbf{Objective function:} In order to train our Generator to synthesise samples as real as possible, we formulate an adversarial loss. More specifically, we use LSGAN \cite{lsgan} with labels $a=c=1$ for fake samples and label $b=0$ for real ones, resulting in the following adversarial objective for the Generator:
\begin{equation}
\begin{split}
& \mathcal{L}_{adv}^G = \dfrac{1}{2} {\rm I\!E}_{t}[(D_D(\mat{W}_{t:t+1}, \tilde{\mat{Y}}_{t:t+2}) - 1)^2] \\
& + \dfrac{1}{2} {\rm I\!E}_{t}[(D_I(\mat X_{t}, \tilde{\mat Y}_{t}) - 1)^2 + (D_M(\mat X_{t}^{m}, \tilde{\mat Y}_{t'}^{m}) - 1)^2].
\end{split}
\label{eq:1}
\end{equation}
We add two more losses in the learning objective function of the Generator: 1) a VGG loss $\mathcal{L}_{vgg}^G$ and 2) a feature matching loss $\mathcal{L}_{feat}^G$, first proposed in the work of Xu et al. \cite{xu2017learning}. Our feature matching loss term is based on the activations of both the Image and Dynamics Discriminators. Given a ground-truth frame $\mat Y_t$ and the synthesised frame $\tilde{\mat Y}_t$, we use the VGG network \cite{vgg} to extract visual features in different layers for both frames and compute the VGG loss as in \cite{pix2pixHD} and \cite{vid2vid}.
Likewise, the feature matching loss is computed by extracting features with the two Discriminators $D_I$ and $D_D$ and computing the $\ell_1$ distance of these features for a fake frame $\tilde{\mat Y}_t$ and the corresponding ground truth $\mat Y_t$. For the computation of the overall feature matching loss, we compute a loss term $\mathcal{L}_{feat}^{G-D_I}$ using features extracted by $D_I$, as well as another loss term $\mathcal{L}_{feat}^{G-D_D}$, using the dynamics discriminator $D_I$ and then we add them: $\mathcal{L}_{feat}^G = \mathcal{L}_{feat}^{G-D_I} + \mathcal{L}_{feat}^{G-D_D}$. The total objective for $G$ is given by:
\begin{equation}
\mathcal{L}^G = \mathcal{L}_{adv}^G + \lambda_{vgg} \mathcal{L}_{vgg}^G + \lambda_{feat}\mathcal{L}_{feat}^G  
\end{equation}
To achieve a balance between the loss terms above, we set $\lambda_{vgg} = \lambda_{feat} = 10$. The Image, Mouth and Dynamics Discriminators are optimised under their corresponding adversarial objective functions (see Supplementary Material).

\textbf{Facial dynamics.} One key factor to consider while designing the Dynamics Discriminator is how it could learn the complex realistic facial muscular interactions observed in monocular videos of facial performances. As discussed and validated experimentally in \cite{Koujan_2020_CVPR}, off-the-shelf state-of-the-art optical flow methods solve this problem without any prior knowledge, since they target any moving objects in the scene. When applied to faces captured in the wild, the estimated flow does not reflect faithfully the non-rigid and composite facial deformations. To tackle this, we employ a state-of-the-art network, termed as FlowNet2, for the optical flow estimation \cite{flownet2}. The authors of \cite{flownet2} trained this network on publicly available images after rendering them with synthesised chairs modified by various affine transformations. Starting from the pretrained models of \cite{flownet2},  we fine-tune their network on the 4DFAB dataset \cite{cheng20184dfab}. This dataset has dynamic high-resolution 4D videos of subjects eliciting spontaneous and posed facial behaviours. Our fine-tuned FlowNet2 network is therefore utilised to estimate the optical flow matrix $\mat{W}$ referred to in equation \ref{eq:1}.

\section{Experiments}

\subsection{Dataset and Implementation Details}

We train and test \textit{Head2Head++} on a newly collected video dataset, which we make publicly available, consisting of eight different individuals. Each individual is depicted in one video, which is at least 10 minutes long and presents the target in diverse head poses and facial expressions.  First, we apply face detection on each video, as a pre-processing step. We utilise \cite{mtcnn} to obtain a bounding box per frame, and then we compute the average bounding box throughout frames. We extract the facial ROI of $256 \times 256$ pixels for each frame, according to this fixed bounding box. This way, the background remains as stable and unchanged as possible, from the beginning of the video until the end. Finally, the frames of each subject are split into a training and a test set. Please refer to the Supplementary Material for more details on the dataset.

We base our Generator's architecture on our previous work \cite{Koujan2020head2head}. All discriminators have the same architecture, adopted by \textit{pip2pixHD} \cite{pix2pixHD}.  
We train a separate person- and video-specific Video Rendering Network for each one of the eight target identities. Each trained model at the end is dedicated for the target video used during training and does not generalise to synthesise the same target person under different cloths and backgrounds (only video specific). Given a target video footage of 10 minutes, extracting the eyes and NMFC conditional input sequences and training, the GAN framework requires around 20 hours. Networks are optimised with Adam \cite{adam}, for 40 epochs with an initial learning rate $\eta = 2 \cdot 10^4$, $\beta_1 = 0.5, \beta_2 =  0.999$, on a single NVIDIA GeForce RTX2080 Ti. During test time, our \textit{Head2Head++} pipeline performs head reenactment from web-camera captures in nearly real-time speeds (18 fps).

\subsection{Supported Reenactment Methods}

\label{subsec:reenmethods}

As already mentioned, we have chosen to use 3DMMs 
for the retrieval of 3D facial shapes from the source and target videos in order to address the target identity preservation problem, allowing us to disentangle expression from identity seamlessly.
Therefore, \emph{Head2Head++} can be used both for full head reenactment and simple facial reenactment, depending on which set of camera parameters (target or source) are used during the creation of the conditional NMFC sequence.

\begin{figure}[t!]
\includegraphics[trim=10 0 0 0, width=3.5in]{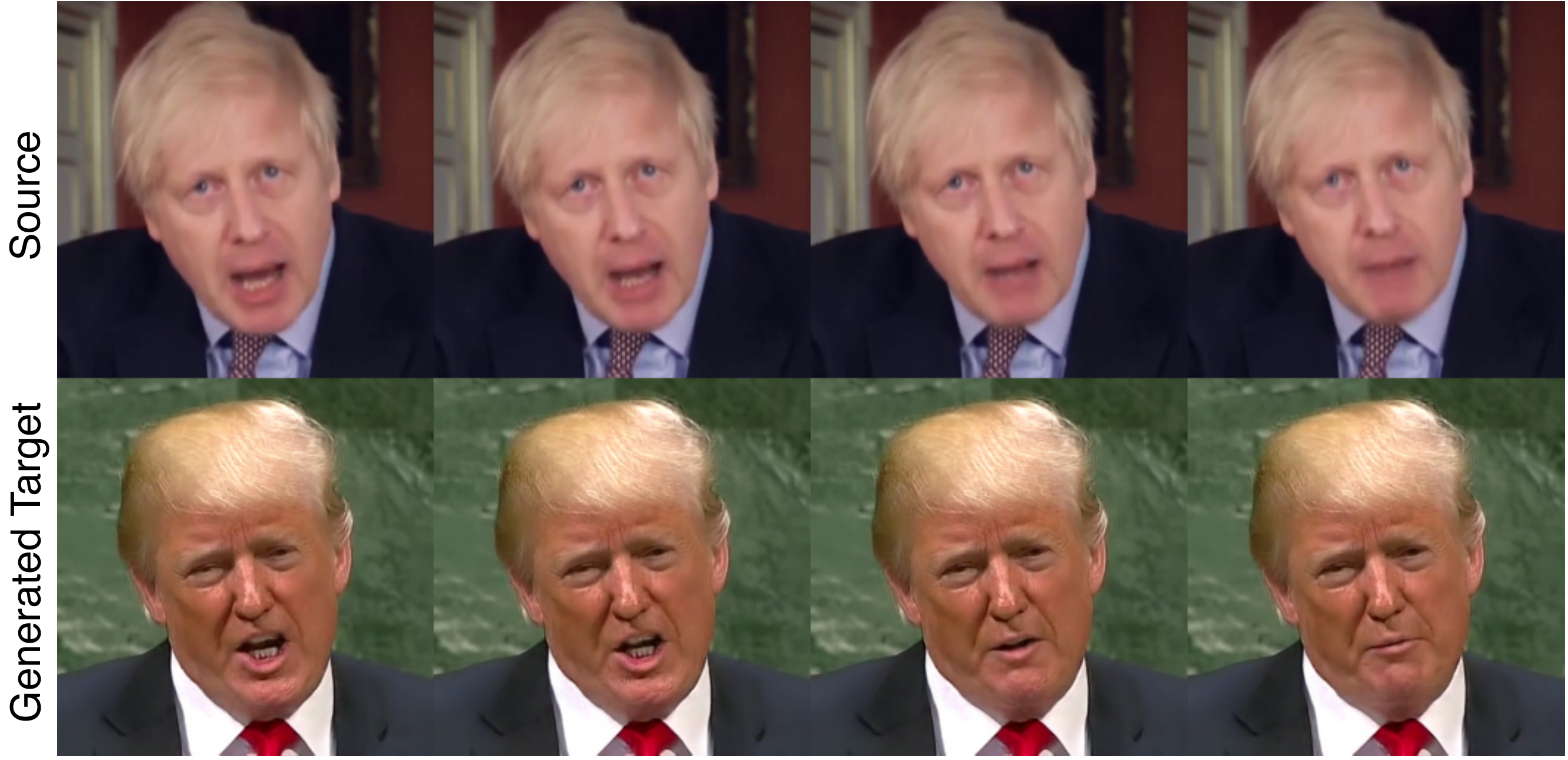}
\caption{Head reenactment aims at transferring the expressions, pose and eye movements from the driving sequence to the target identity.}
\label{fig2}
\end{figure}

\begin{figure}[t!]
\includegraphics[trim=24 0 0 0, width=3.5in]{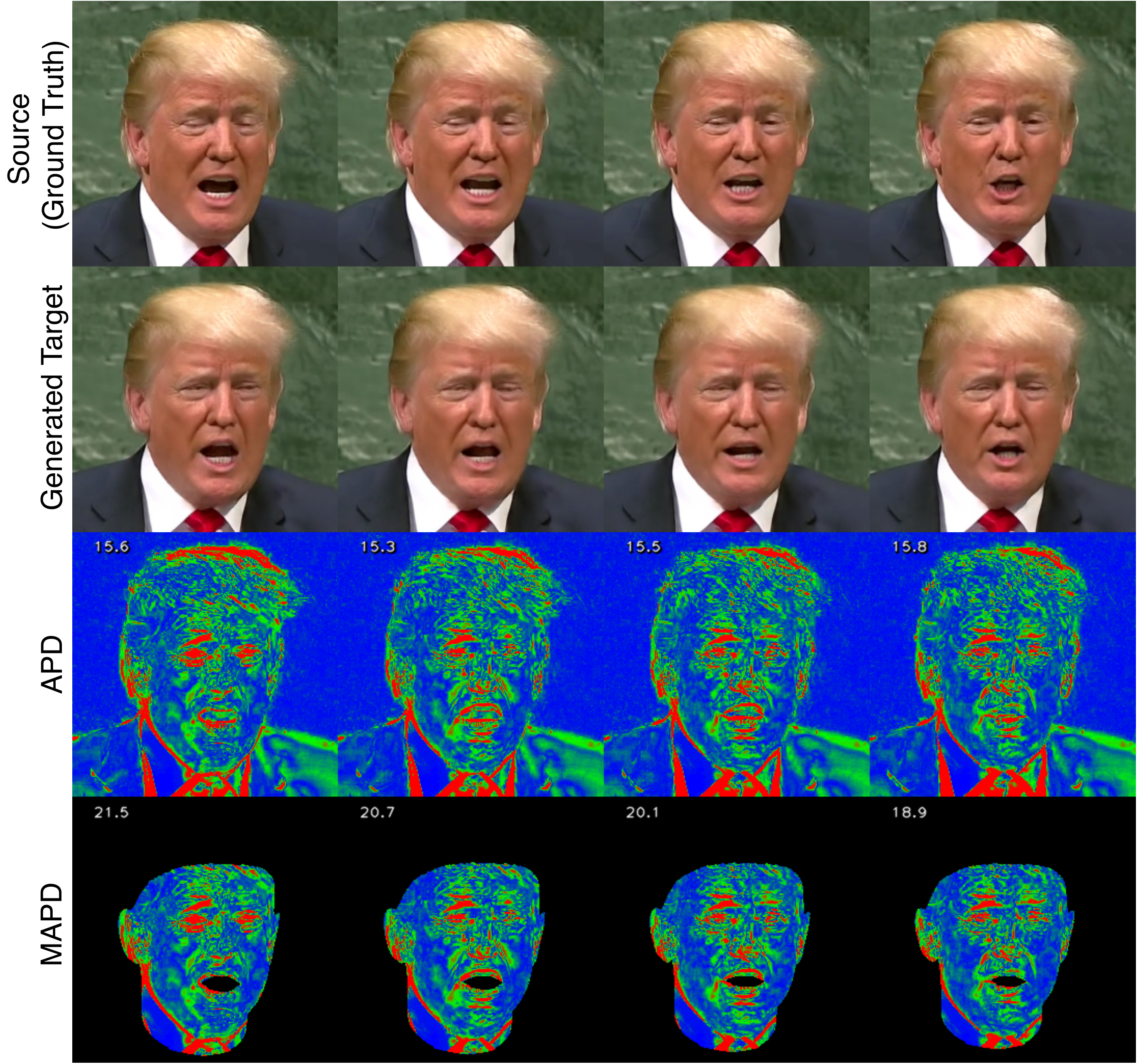}
\caption{Self reenactment is used for evaluating the performance of models. Here, the source identity coincides with the target one. We display Average Pixel Distance (APD) and Masked Average Pixel Distance (MAPD) between the generated and ground truth frames, in the form of RGB heatmaps. Please refer to Sec.~\ref{subsec:reenmethods} for more details on those metrics.}
\label{fig3}
\end{figure}


\textbf{Head Reenactmenent.} This is the main functionality of our method and arguably the most challenging, since it involves transferring the complete head motion from any driving video to the desired target subject. A head reenactment example is demonstrated in Fig. \ref{fig2}. Given the driving sequence (Johnson), our method synthesises a highly photo-realistic and temporally coherent video of the targeted identity (Trump). 


\textbf{Face Reenactmenent.} Instead of transferring all head movements from the source video, we can simply pass the facial expressions to the target identity. This is done by utilizing the original camera parameters, estimated from the target training sequence. In this case, we use the original eye landmarks extracted from the training sequence to drive synthesis. The advantage of face reenactment is that our Generator has encountered the same head poses and positions during training, which is beneficial when it comes to synthesising the hair, background and upper body areas, where no conditional information is available in the input. Although prior works have mainly used it as a manipulation method \cite{face2face, DeferredNeuralRendering}, we perform face reenactment by generating all pixels within frames, not only a masked area of the face.

\textbf{Self Reenactmenent.} During self reenactment, the source identity coincides with the target one. This is extremely useful for training and evaluating our model, since we are given access to the ground truth frames. Note that the driving sequences used during test have not been seen during training, since they belong to the test data split. Fig. \ref{fig3} shows a self reenactment experiment. Ideally, the generated frames should be identical with the source ones. For the evaluation of the reconstructive ability of models in the experiments that follow, we use metrics such as the average pixel distance (APD) or masked average pixel distance (MAPD) between the synthesised and ground truth frames. In Sec.~\ref{subsec:metrics}, we provide more details on APD and MAPD. 


\subsection{Quantitative Evaluation Metrics}
\label{subsec:metrics}

We evaluate the performance of our system both qualitatively and quantitatively. Most experiments were performed under self reenactment. That is, given a set of test frames $\mat Y^{(i)}$ for each target identity $i$ in the dataset, we generated the corresponding synthetic frames $\tilde{\mat Y}^{(i)}$ and used $\mat Y^{(i)}$ as ground truth. We assessed our method's reconstructive ability, target identity preservation, pose and expression transferability as well as the photorealism of generated frames, using the metrics below:

 \textbf{Average Pixel Distance (APD):} is computed as the average L2-distance of RGB values across all spatial locations and frames, between the ground truth and generated data.
 
 \textbf{Masked Average Pixel Distance (MAPD):} similar to APD, it tests the reconstructive performance. A mask computed from NMFC frames is used to constrain the metric on the facial area, where conditional information is available. 
 
 \textbf{Distance between Average Identities (DAI):} after performing 3D face reconstruction on $\mat Y^{(i)}$ and $\tilde{\mat Y}^{(i)}$ sequences, we compute the average identity coefficients for each sequence, and then the L1-distance between those average identities.
 
 \textbf{Average Expression Distance (AED):} after reconstructing the fake and ground truth sequences, we measure the average L1-distance between the expression coefficients across frames.
 
 \textbf{Average Rotation Distance (ARD)}: this metric is used to measure pose transferability. Using the estimated camera parameters from the generated and ground videos, we compute the Euler angles that correspond to head poses. Then, the average Euler angles discrepancy across sequences is determined in terms of degrees.
 
 \textbf{Fréchet Inception Distance (FID):} this is a widely used metric for evaluating the visual quality of individual frames, originally proposed in \cite{NIPS2017_7240}. We use \cite{Deng_2019_CVPR} as a feature extractor and we report the mean FID throughout the experiments.
 
 \textbf{Maximum Mean Discrepancy (MMD$^2$):} is a very useful metric for measuring the discrepancy between real and fake frames \cite{mmd}.

\subsection{Ablation Study}

\begin{figure}[t!]
    \centering
    \includegraphics[width=3.5in]{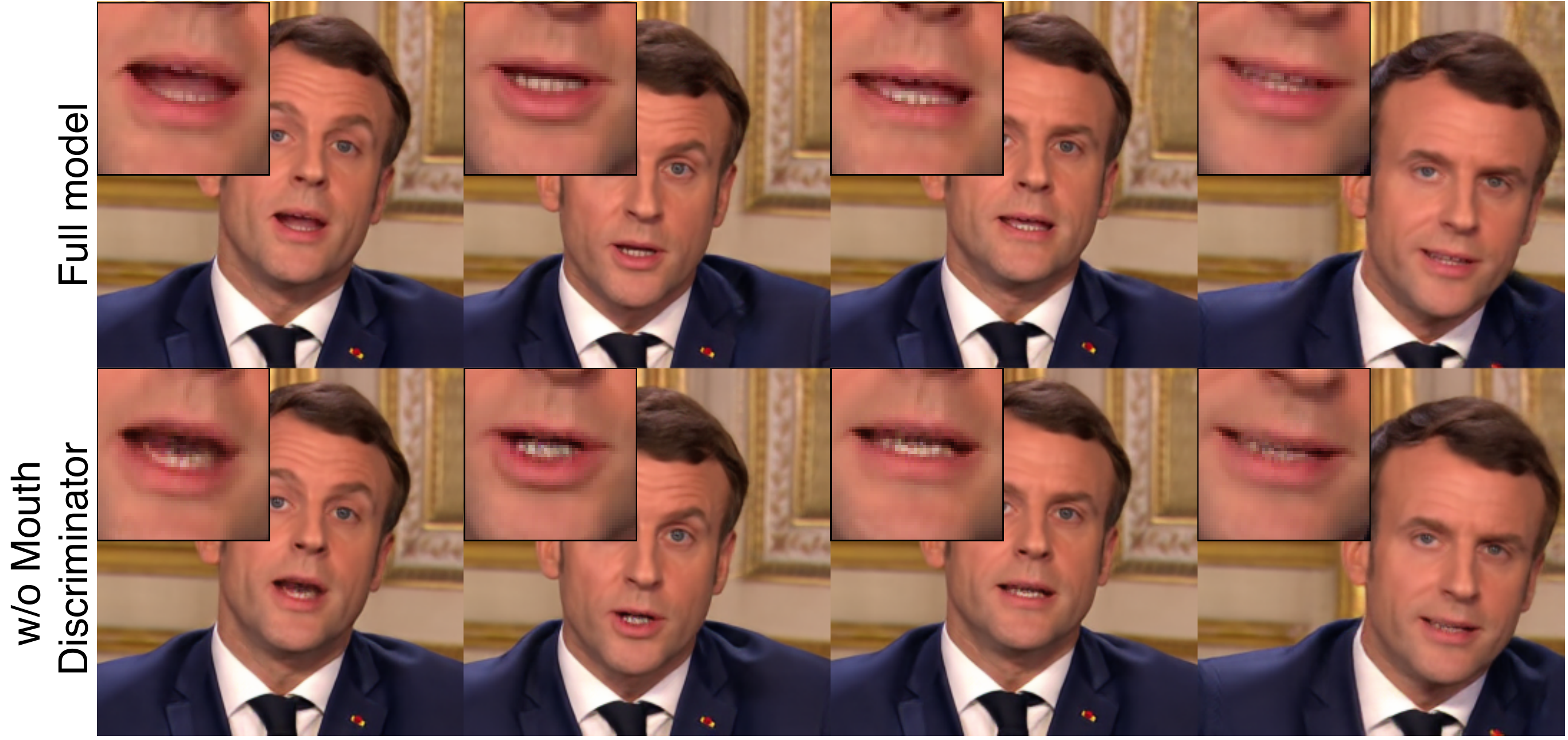}
    \caption{Significance of Mouth Discriminator. Please zoom in for details.}
    \label{fig:fig5}
\end{figure}

\begin{figure}[t!]
    \centering
    \includegraphics[width=3.5in]{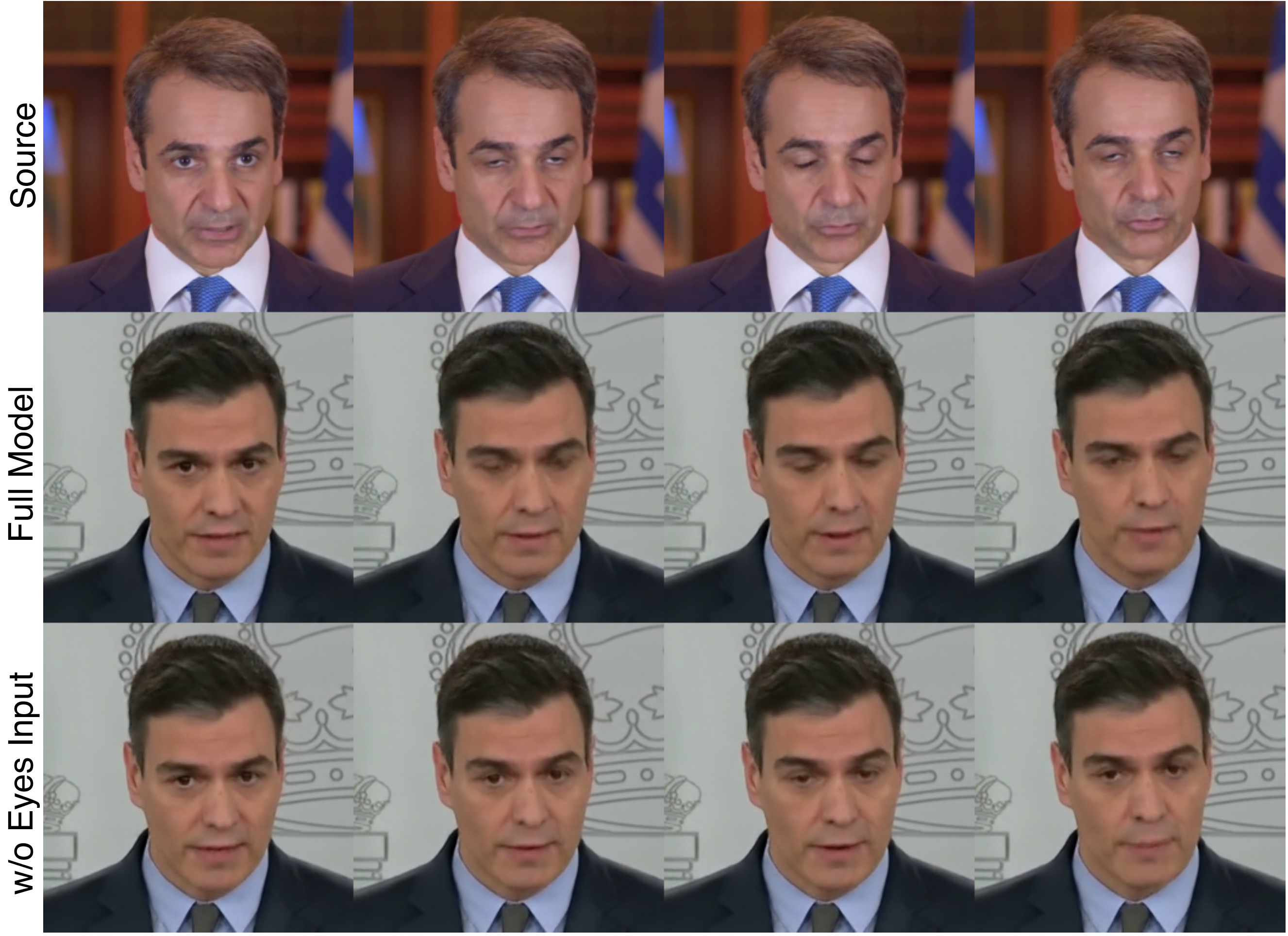}
    \caption{Significance of conditioning on the eyes. The blinking of the source is successfully transferred to the target when using eye landmarks to condition synthesis. On the contrary, conditioning solely on NMFC frames is not sufficient for transferring eye gaze and blinking.}
    \label{fig:fig4}
\end{figure}

We conducted an ablation study, in order to assess the significance of several "key" components of our \textit{Head2Head++} system. First, we evaluate numerically the contribution of the two reconstructive loss terms, namely the feature matching and VGG loss. As can be seen in the first two rows of Table \ref{table1}, all metrics demonstrate the contribution of these loss terms on the quality of the generated samples. 
The next row of the table validates 
the importance of a sequential Generator, as opposed to its non-sequential variation (``w/o seq.~$G$''). The results indicate that considering previously generated frames in the Generator has a very positive effect on the visual quality of samples. Finally, the majority of metrics show that removing the Dynamics Discriminator (``w/o $D_D$'') might slightly improve the quality of frames, when seen as individual images. We believe this is due to the fact that the proposed metrics analyse solely the spatial content of frames and do not evaluate temporal information. Removing $D_D$ network would actually reduce the temporal stability and coherence of videos. Such behaviour is more apparent in areas where no conditional information is available (e.g. hair, background, upper body). In order to understand better the significance of the Dynamics Discriminator in the visual plausibility of the output video, please refer to our Supplementary Video.

\begin{table}[!t]
\renewcommand{\arraystretch}{1.3}
\caption{Ablation study results under the self reenactment setting, averaged across all eight target identities in our dataset. For all metrics, lower values indicate better quality or performance. Bold and underlined values correspond to the best and the second-best
value of each metric, respectively.}
\label{table1}
\centering
\begin{tabular}{|c|c|c|c|c|c|c|}
\hline
 &  &  &  &  & $\times 10^2$ & $\times 10^5$ \\
\textbf{Variations} & APD & MAPD & AED & ARD & FID & MMD$^2$ \\
\hline
\hline
w/o $\mathcal{L}_{feat}^G$ & 17.62 & 16.70 & 0.627 & $0.539^\circ$ & 6.98 & 13.23 \\
\hline
w/o $\mathcal{L}_{vgg}^G$ & 15.74 & 13.95 & 0.514 & $0.452^\circ$ & 5.40 & 9.42 \\
\hline
w/o seq. $G$ & 15.71 &13.63 & 0.486 & $0.436^\circ$ & 4.98 & 9.06 \\
\hline
w/o $D_D$ & \underline{15.09} & \textbf{13.03} & \textbf{0.456} & $\textbf{0.408}^\circ$ & \textbf{3.95} & \textbf{5.57} \\
\hline
\textbf{\textit{Full model}} & \textbf{14.82} & \underline{13.06} & \underline{0.467} & $\underline{0.412}^\circ$ & \underline{4.15} & \underline{6.50} \\
\hline
\end{tabular}
\end{table}



 \begin{table}[h]
 \parbox{.45\linewidth}{
 \centering
 \caption{Ablation study of the effect of Mouth Discriminator $D_M$ and conditioning on the eyes in \textit{Head2Head++}.}
 \label{tab:EvalMouthEye}
 \begin{tabular}{|c|c|c|}
 \hline
 \textbf{Variations} & APD & AELD \\
 \hline
 \hline
 w/o $D_M$ & 13.43 & - \\
 \hline
 w/o Eyes input & - & 1.52 \\
 \hline
 \textbf{\textit{Full model}} & \textbf{12.32} & \textbf{1.06} \\
 \hline
 \end{tabular}
 }
 \hfill
 \parbox{.45\linewidth}{
 \centering
 \caption{Ablation study of the effect of Sequential Generator and Dynamics Discriminator in \textit{Head2Head++}.}
 \label{tab:EvalDynamics}
 \begin{tabular}{|c|c|}
 \hline
 \textbf{Variations} & FVD \\
 \hline
 \hline
 w/o seq. $G$ & 74.42 \\
 \hline
 w/o $D_D$ & 66.48 \\
 \hline
 \textbf{\textit{Full model}} & \textbf{57.46} \\
 \hline
 \end{tabular}
 }
 \end{table}

Arguably, the eyes and the mouth, are the facial areas that might be the first to expose the "fakeness" of a synthetic video. In Fig. \ref{fig:fig5}, we illustrate the importance of the dedicated Mouth Discriminator, initially proposed in \cite{Koujan2020head2head}. As can be seen in the second row of Fig. \ref{fig:fig5}, when the Mouth Discriminator is not included in GAN training, teeth details make the generated video appear less realistic, as opposed to the results in the first row, where the Generator was supervised by $D_M$ network. Fig. \ref{fig:fig4} demonstrates the significance of conditioning the Generator on the eyes input sequence. Given that NMFCs do not adequately reflect eye gaze and blinking, conditioning solely on the NMFC input sequence does not provide substantial information to the Generator, for transferring the eye movements of the source to the target. Additionally, we quantitatively evaluate the significance of the mouth discriminator and the eye gaze conditioning. Table \ref{tab:EvalMouthEye} presents the numeric measures obtained when assessing their effect. In the second column of Table \ref{tab:EvalMouthEye}, we report the \textbf{average pixel distance (APD)} within the\textbf{ mouth area} under self-reenactment, between the generated and ground truth regions of interest, for all eight identities in our dataset. As can be seen, the numeric results agree with the qualitative ones, in Fig. 4. In the third column of Table \ref{tab:EvalMouthEye}, we assess the importance  of \textbf{conditioning the Generator on the Eyes input} (Eyes Video sketch shown in Fig. 1). For that, we use the landmarker to extract eye landmarks, both from the ground truth and synthetic frames, created under self-reenactment. Then, we compute the \textbf{average eye landmarks $L2$-distance (AELD)}, between real and generated frames, across all eight identities and report the results in the Table \ref{tab:EvalMouthEye}. We observe that conditioning on the eyes helps to synthesise faces that better follow the eye movement and gaze appearing in the source video.

Lastly, in order to evaluate the importance of the Dynamics Discriminator $D_D$ on the temporal consistency of generated frames, we use the Fréchet Video Distance (FVD) \cite{fvd}. We also provide the FVD score for the non-sequential variation of Head2Head++ (w/o seq. $G$). The results of Table \ref{tab:EvalDynamics} indicate that our full model with a sequential Generator trained alongside a Dynamics Discriminator outperforms both variations by a significant margin.

\subsection{Effect of Training Video Length}

Fig. \ref{fig:fig13} displays the influence of training footage duration on the generative performance of the the Video Rendering Network. For that experiment, we trained five separate models per target identity. Each model was trained on frame sequences of different length: one, two, four, eight, up to sixteen thousand frames. We found that FID and MMD$^2$ scores, which indicate visual quality, are the metrics affected the most by the number of training frames. Furthermore, it is visible that there is still room for improvement in the performance of our Video Rendering Network, provided that even longer training videos are available. However, this would require additional footage with enough head pose variability, which in practice does not scale linearly with the number of training frames available.

\begin{figure}[t!]
    \centering
    \includegraphics[width=3.5in]{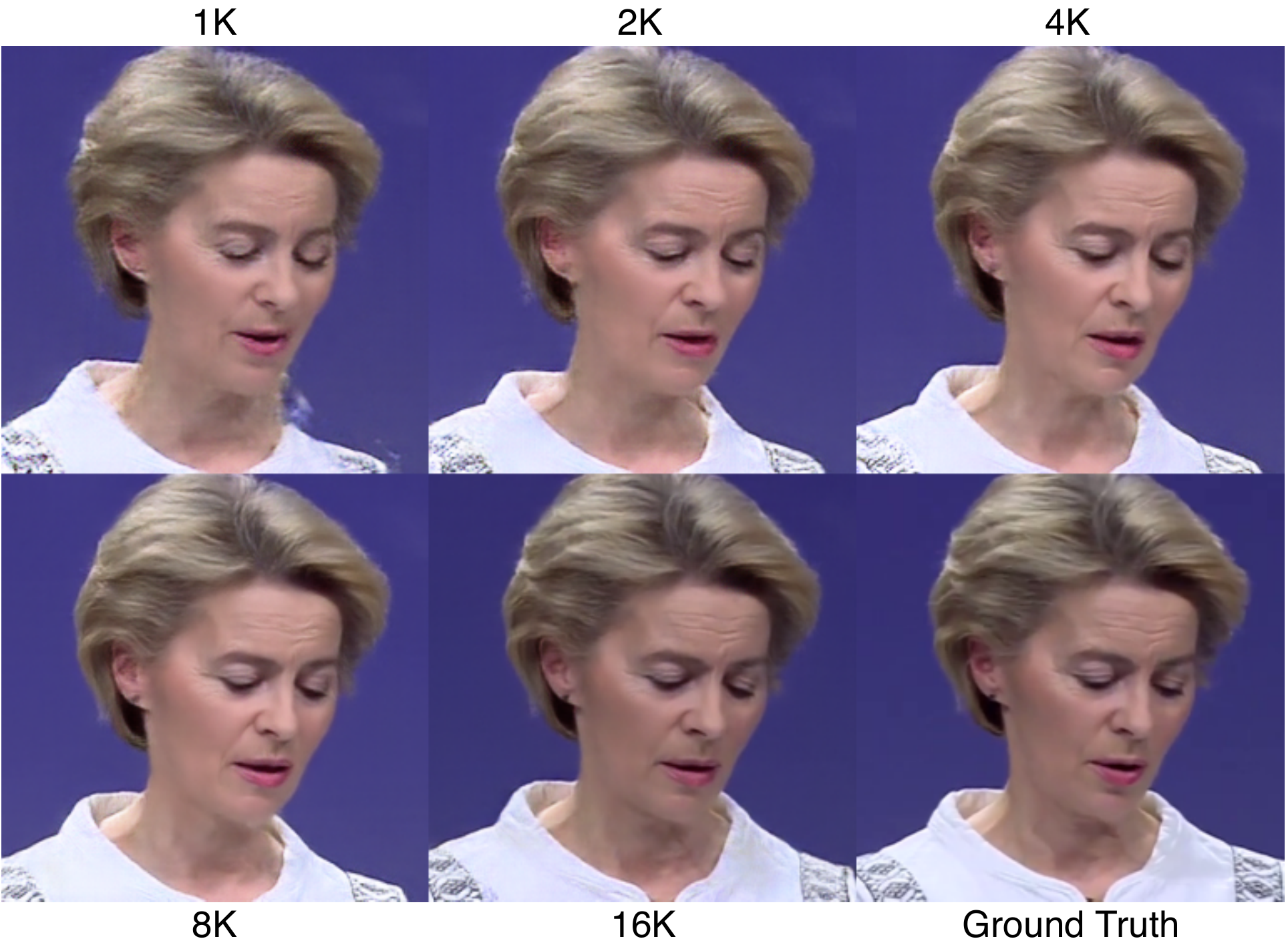}
    \caption{Significance of training video length. We show the same image generated with a GAN trained on 1K, 2K, 4K, 8K and 16K frames, as well as the ground truth (last image).}
    \label{fig:fig7}
\end{figure}

\begin{figure}[h!]
\centering
\pgfplotsset{every tick label/.append style={font=\huge \boldmath}}
\pgfplotsset{xlabel style={align=center}}
\subfloat[]{
\begin{tikzpicture}[scale=0.34]
\begin{axis}[
    xlabel={\Huge number of frames \\ \Huge (in thousands)},
    xmin=0, xmax=17,
    ymin=10, ymax=30,
    xtick={1,2,4,8,16},
    ytick={14,24},
    ymajorgrids=true,
    legend pos=north east,
    grid style=dashed,
]
\addplot[
    color=blue,
    mark=*,
    ]
    coordinates {
    (1,24.27)(2,22.29)(4,21.97)(8,20.57)(16,17.16)
    };
\addplot[
    color=red,
    mark=*,
    ]
    coordinates {
    (1,21.99)(2,19.80)(4,18.95)(8,18.35)(16,14.03)
    };
\legend{\Huge APD, \Huge MAPD}
\end{axis}
\end{tikzpicture}
}
\subfloat[]{
\begin{tikzpicture}[scale=0.34]
\begin{axis}[
    xlabel={\Huge number of frames \\ \Huge (in thousands)},
    xmin=0, xmax=17,
    ymin=0.3, ymax=1.4,
    xtick={1,2,4,8,16},
    ytick={0.4,1.1},
    ymajorgrids=true,
    legend pos=north east,
    grid style=dashed,
]
\addplot[
    color=blue,
    mark=*,
    ]
    coordinates {
    (1,0.838)(2,0.682)(4,0.579)(8,0.513)(16,0.453)
    };
\addplot[
    color=red,
    mark=*,
    ]
    coordinates {
    (1,1.060)(2,0.772)(4,0.601)(8,0.572)(16,0.461)
    };
\legend{\Huge AED, \Huge ARD}
\end{axis}
\end{tikzpicture}
}
\subfloat[]{
\begin{tikzpicture}[scale=0.34]
\begin{axis}[
    xlabel={\Huge number of frames \\ \Huge (in thousands)},
    xmin=0, xmax=17,
    ymin=0, ymax=45,
    xtick={1,2,4,8,16},
    ytick={4,32.2},
    ymajorgrids=true,
    legend pos=north east,
    grid style=dashed,
]
\addplot[
    color=blue,
    mark=*,
    ]
    coordinates {
    (1,15.19)(2,9.75)(4,6.64)(8,5.17)(16,3.94)
    };
\addplot[
    color=red,
    mark=*,
    ]
    coordinates {
    (1,32.26)(2,17.39)(4,10.85)(8,7.57)(16,4.62)
    };
\legend{\Huge FID, \Huge MMD$\mathbf{^2}$}
\end{axis}
\end{tikzpicture}
}
\caption{The effect of the number of training frames on the quality of generated frames. All metrics were computed for the subset of four target identities (Macron, Sanchez, Trump, Von Der Leyen).}
\label{fig:fig13}
\end{figure}
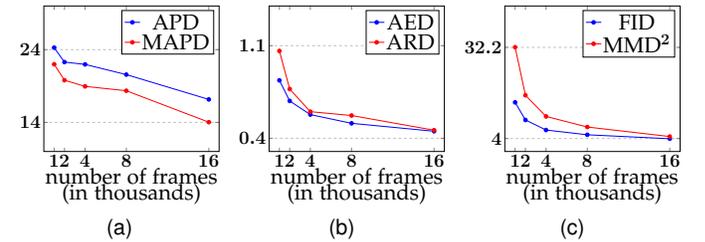


In Fig. \ref{fig:fig7}, we demonstrate the impact of training video length on the visual quality of synthetic frames. We observe visually that as the number of available training frames increases, artifacts become less severe and eventually disappear. This behaviour comes in agreement with the decrease of evaluation metrics, shown in Fig. \ref{fig:fig13}.

\begin{figure}[t!]
    \centering
    \includegraphics[width=3.5in]{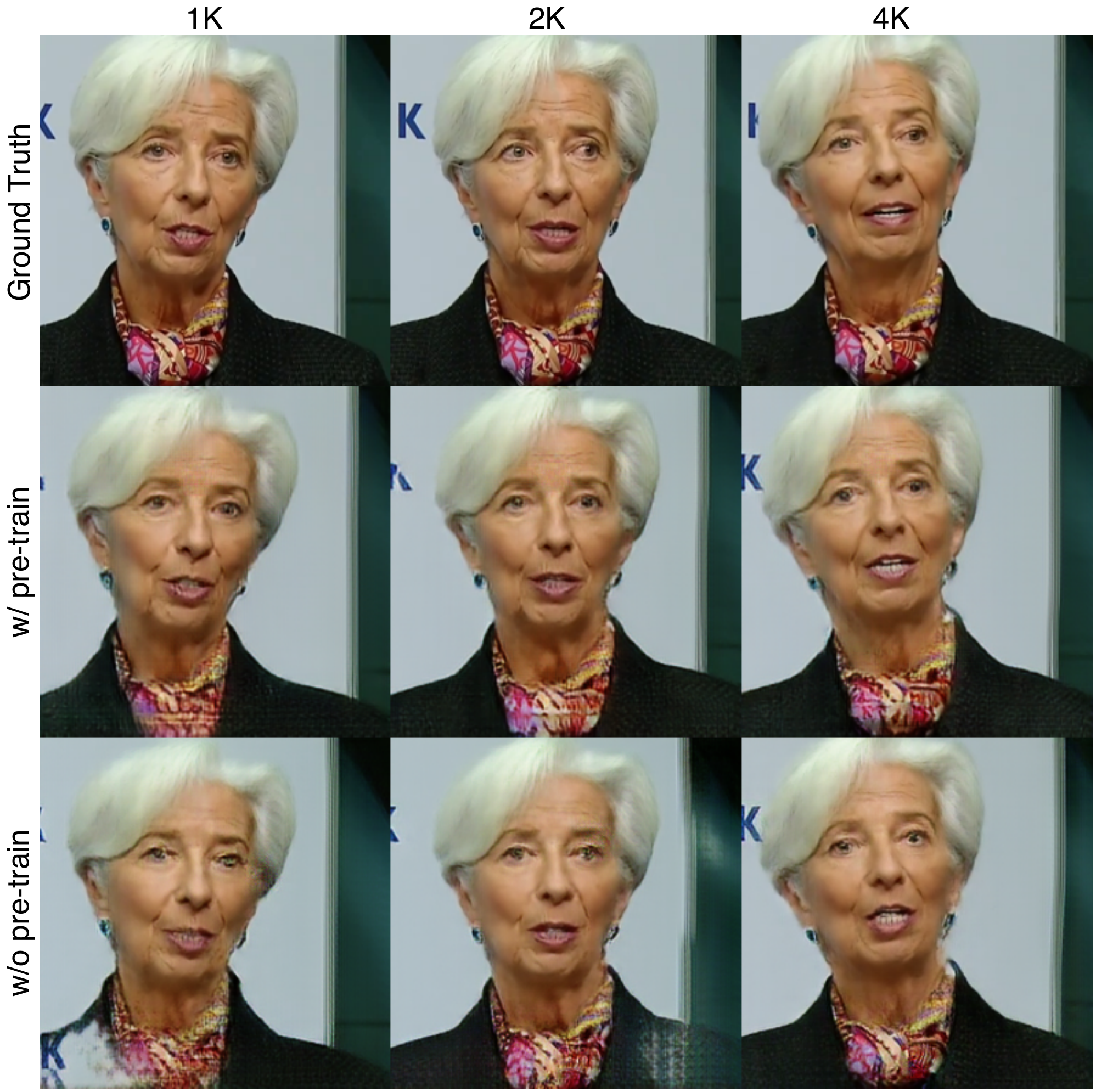}
    \caption{The effect of pre-training the Video Rendering Network on Face Forensics++ dataset. The importance of pre-training is very prominent when a limited number of training frames (1K, 2K) is available.}
    \label{fig:fig8}
\end{figure}

\subsection{Pre-training on Face Forensics++ Dataset}

We further explore how the generative performance of the Video Rendering Network can be improved in the case of limited video footage, for instance  when training sequences are less than four thousand frames (about 2.5 minutes). To that end, we pre-trained \textit{Head2Head++} on the large-scale \textit{Face Forensics++} dataset  \cite{roessler2019faceforensicspp}, containing 1000 different identities. At this stage, the Generator learned to create a ``average identity'', since there is no mechanism to dictate which identity to be synthesised over this multi-person dataset. As a next step, we fine-tuned eight person-specific models, one for each target identity in our dataset, using a subset of their available footage (1K, 2K or 4K frames). In Table \ref{table3}, we report a quantitative evaluation with and without pre-training on \cite{roessler2019faceforensicspp} data. As suggested by the results, pre-training on \textit{Face Forensics++} data appears advantageous, especially for shorter sequences (1K or 2K frames). This could be explained if we consider pre-training as a head start for the GAN, where the Generator is already capable of creating a realistic ``average identity'' and then it remains to learn a specific one. The same trend can be seen in Fig. \ref{fig:fig8}. Especially for a training footage of 1K or 2K frames, the quality of generated frames is clearly inferior in comparison with samples created with pre-training. 

\begin{table}[t!]
\renewcommand{\arraystretch}{1.3}
\caption{Effect of pre-training GAN on Face Forensics++ dataset, when training video footage is limited, namely less than 4K frames. All metrics are averaged across all eight target identities in our dataset. For all metrics, lower values indicate better quality or performance.}
\label{table3}
\centering
\begin{tabular}{|c|c|c|c|c|c|c|}
\hline
\textbf{Number of} & &  &  & & $\times 10^2$ & $\times 10^5$ \\
\textbf{train frames} & APD & MAPD & AED & ARD & FID & MMD$^2$ \\
\hline
\hline
1K & 22.62 & 20.32 & 0.844 & $0.955^\circ$ & 16.01 & 39.59 \\
\hline
w/ pre-train & \textbf{21.95} & \textbf{19.02} & \textbf{0.575} & $\textbf{0.562}^\circ$ & \textbf{15.93} & 39.76 \\
\hline
\hline
2K & 20.41 & 18.27 & 0.719 & $0.737^\circ$ & 11.04 & 24.83 \\
\hline
w/ pre-train & \textbf{20.17} & \textbf{17.84} & \textbf{0.526} & $\textbf{0.516}^\circ$ & 11.22 & 24.93 \\
\hline
\hline
4K & 18.87 & 17.46 & 0.610 & $0.577^\circ$ & 7.41 & 15.08 \\
\hline
w/ pre-train & 18.90 & \textbf{16.71} & \textbf{0.483} & $\textbf{0.453}^\circ$ & 7.80 & 15.63 \\ 
\hline
\end{tabular}
\end{table}

\subsection{Comparison with the State of the Art}

\begin{figure}[t!]
    \centering
    \includegraphics[width=3.5in]{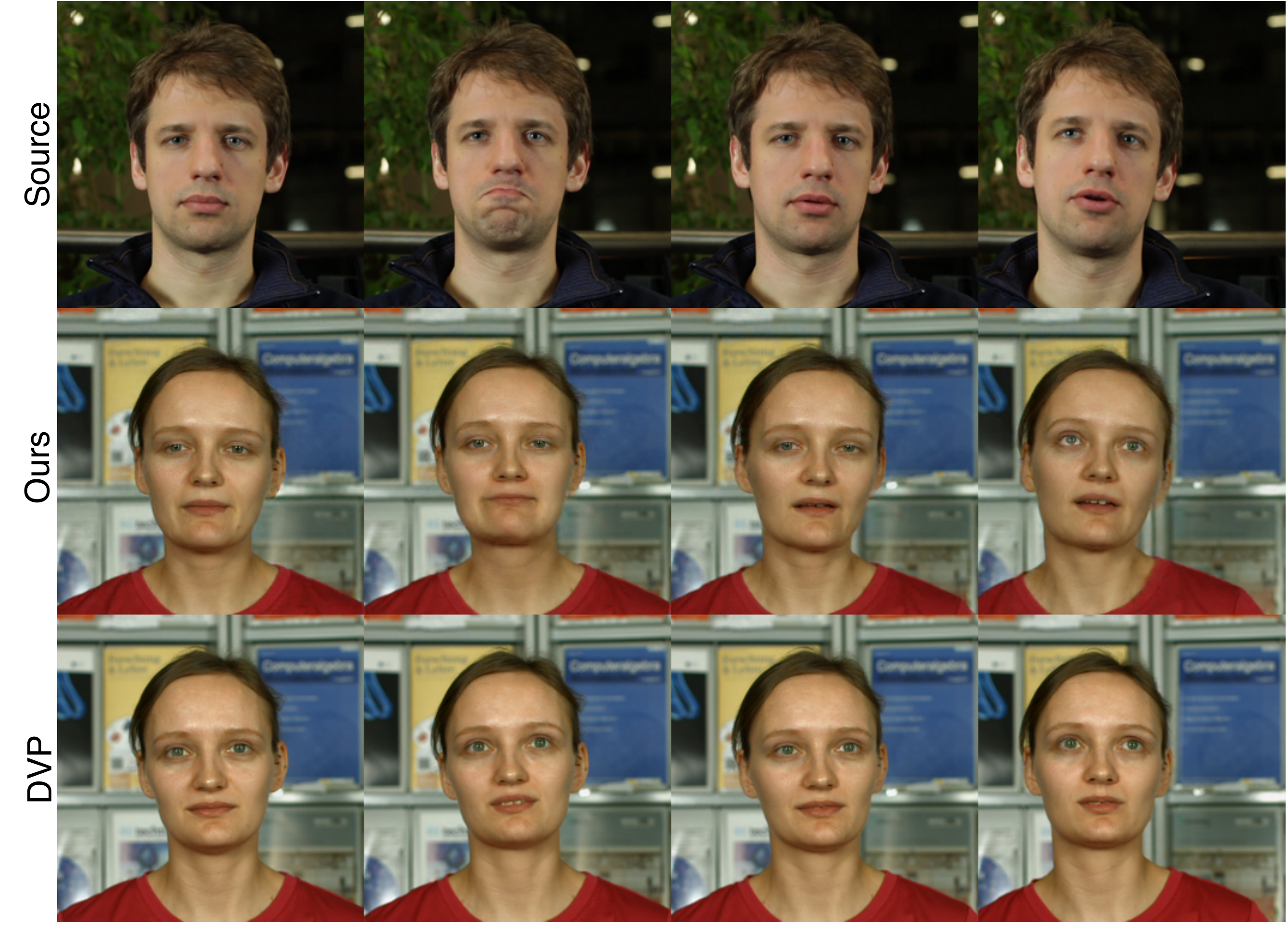}
    \caption{Comparison of our method with \textit{DVP}. In many cases our method outperforms \textit{DVP}, especially in terms of head pose transferability.}
    \label{fig:fig9}
\end{figure}

We compare our \textit{Head2Head++} system with the image-based method of \textit{DVP} \cite{deepvideoportraits} and video-based \textit{Head2Head} \cite{Koujan2020head2head}. This is done by performing a full head reenactment experiment on the same source and target video footage. First, we trained our method and \textit{Head2Head} on the target sequence, which was kindly provided by the authors of \cite{deepvideoportraits}, and then we used the same source to drive synthesis. In Fig. \ref{fig:fig9} we display some examples, in which our method outperforms \textit{DVP} in terms of head pose and expression transferability.

\begin{table}[t!]
\renewcommand{\arraystretch}{1.3}
\caption{Quantitative comparison with \textit{DVP} and \textit{Head2Head} under full head reenactment. For all metrics, lower is better.}
\label{table4}
\centering
\begin{tabular}{|c|c|c|c|c|c|c|}
\hline 
 &  &  &  & $\times 10^2$ & $\times 10^5$ &  \\
\textbf{Method} & AED & ARD & DAI & FID & MMD$^2$ & AELD \\
\hline
\textit{DVP} & 2.089 & $2.54^\circ$ & 23.01 & 15.94 & 61.32 & 0.85 \\
\hline
\textit{Head2Head} & \textbf{1.426} & $\textbf{0.70}^\circ$ & 9.56 & 39.35 & 186.11 & 0.72 \\
\hline
\textbf{\textit{Ours}} & 1.509 & $0.83^\circ$ & \textbf{8.86} & \textbf{12.42} & \textbf{39.15} & \textbf{0.71} \\
\hline
\end{tabular}
\end{table}

For a quantitative comparison of the three methods, we performed 3D face reconstruction on the two synthetic videos and computed the average expression distance (AED) and average rotation distance (ARD) across frames, between the source video and the three generated videos. Then, we applied 3D face reconstruction on the target video and computed the distance between the average identities extracted from this real and each one of the three fake sequences. Finally, we used the FID and MMD$^2$ scores as a photo-realism metric, both computed between the fake videos and a set of frames from the target's original footage. For the eye region, where significant visual improvements have been made over \textit{Head2Head}, we computed the average eye landmarks $L2$-distance (AELD), between source and generated videos.

\begin{figure}[t!]
    \centering
    \includegraphics[width=3.5in]{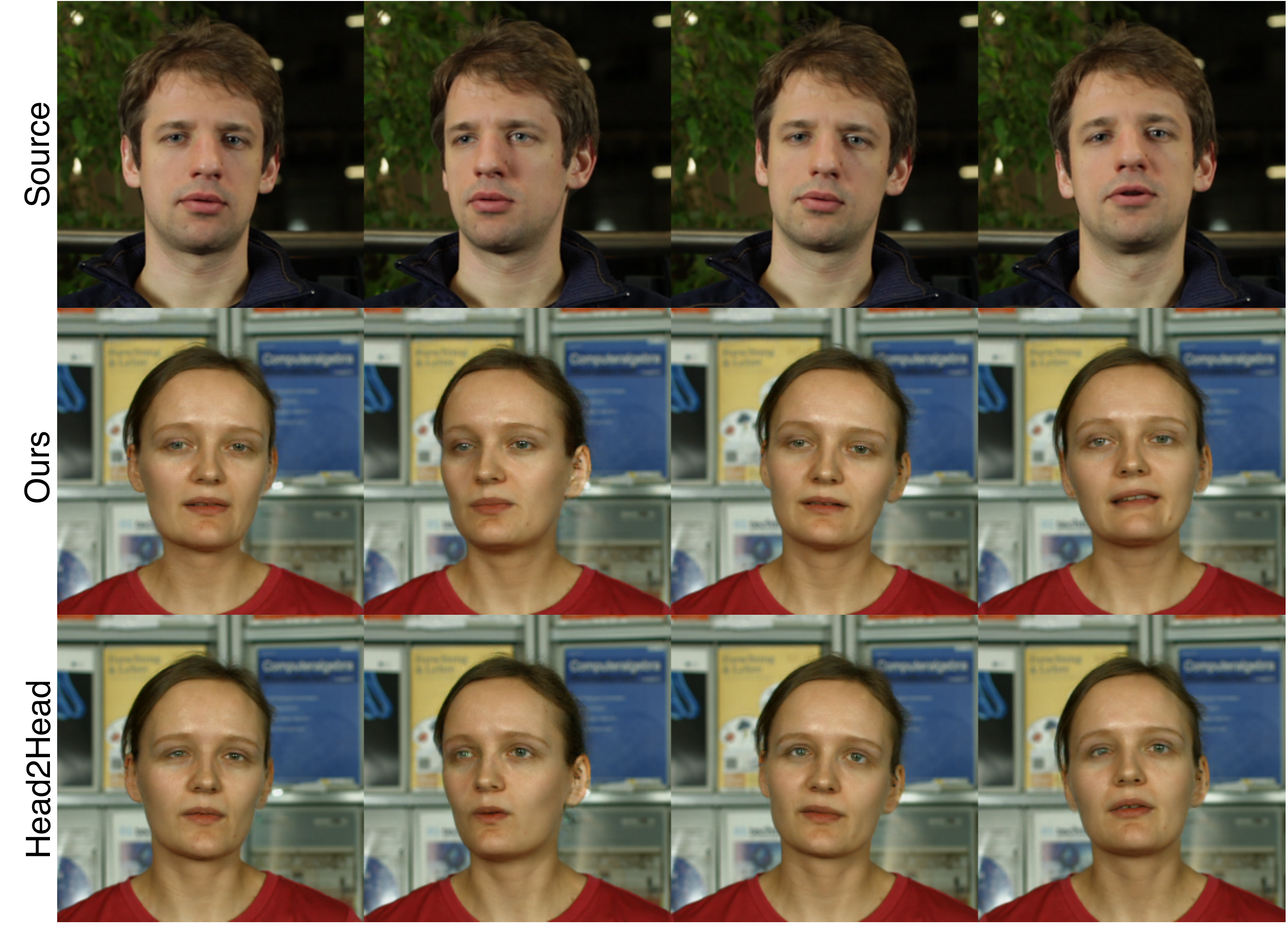}
    \caption{Comparison of our \textit{Head2Head++} method with \textit{Head2Head} \cite{Koujan2020head2head} on eye region synthesis. We synthesise more photo-realistic eyes in comparison to \cite{Koujan2020head2head}. Please refer to the supplementary video for more examples.}
    \label{fig:fig10}
\end{figure}

The results presented in Table \ref{table4} suggest that our method outperforms DVP on every single metric and \textit{Head2Head} in several. More specifically, data samples produced by \textit{Head2Head++} exhibit significantly lower FID and MMD$^2$ scores, when compared to these created by \textit{Head2Head++}. This can be explained by the fact that the  network (ArcFace \cite{Deng_2019_CVPR}) we employ to extract facial features for FID and MMD$^2$ is very sensitive to the shape of the eyes and thus imperfections and inconsistencies in the eyes region generated by \textit{Head2Head} are reflected in both photo-realism metrics. Considering the AELD metric, the difference between the two methods is very small, which can be justified by the fact that the eye tracker used in \textit{Head2Head} is quite reliable and the video mostly frontal, therefore differences with \textit{Head2Head++} can be better understood in terms of photo-realism. In Fig. \ref{fig:fig10}, we show that our new eye motion extraction method results in more reliable eye synthesis, compared to \textit{Head2Head} \cite{Koujan2020head2head}, in terms of photo-realism and eye shape.  We observe that the eye regions generated by \textit{Head2Head} seem unnatural, while \textit{Head2Head++} has successfully rendered photo-realistic eyes.

The AED, ARD and DAI metric values obtained for \textit{Head2Head} seem to be slightly better than \textit{Head2Head++}. We attribute these results to the fact that \textit{Head2Head} is equipped with a video-based 3D reconstruction approach that relies on all video frames to generate the 3D faces, taking a few minutes compared to few milliseconds in the case of \textit{Head2Head++}. Nonetheless, we strongly believe that performing 3D face recovery in nearly real-time outweighs the small disadvantage we noticed in terms of facial expression accuracy.

\begin{table}[t!]
\renewcommand{\arraystretch}{1.3}
\caption{Quantitative comparison with \textit{X2Face} and \textit{FOMM}, using all eight target identities in our dataset, under self reenactment.}
\label{table5}
\centering
\begin{tabular}{|c|c|c|c|}
\hline
\textbf{Method} & APD & FID ($\times 10^2$) & MMD$^2$ ($\times 10^5$) \\
\hline
\textit{X2Face} \cite{X2Face} & 30.54 & 63.3 & 253.2 \\
\hline
\textit{FOMM} \cite{Siarohin_2019_NeurIPS} &  22.43 & 24.6 & 90.3 \\
\hline
\textbf{\textit{Ours}} & 14.82 & 4.2 & 6.5 \\
\hline
\end{tabular}
\end{table}

\begin{figure}[t!]
    \centering
    \includegraphics[width=3.5in]{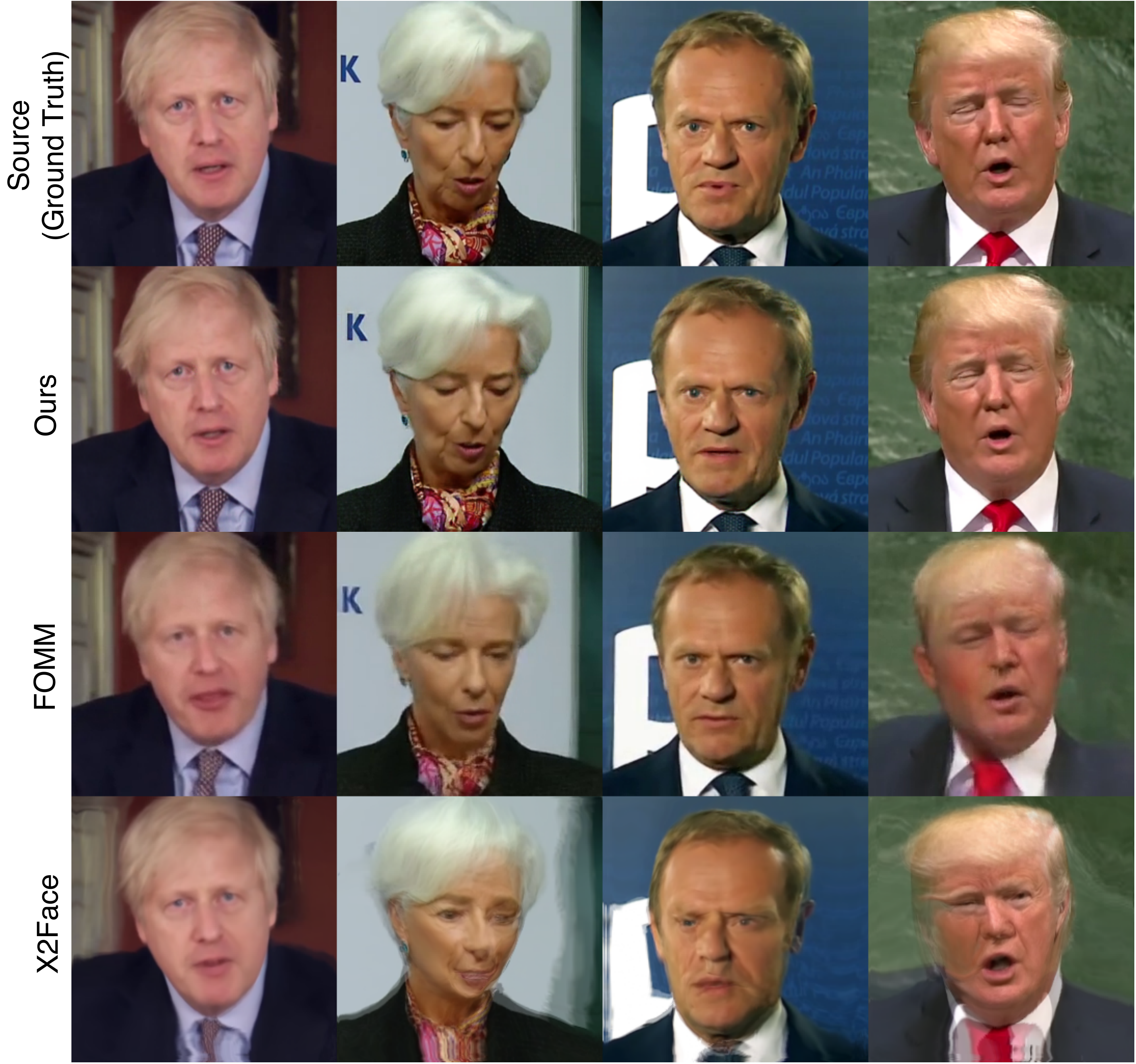}
    \caption{Comparison with \textit{X2Face} and \textit{FOMM}. Our method exhibits better photo-realism. In many cases, the results of \textit{X2Face} appear distorted, while \textit{FOMM}'s performance drops significantly for poses distant from the one provided in the target image.}
    \label{fig:fig11}
\end{figure}

We further compare our method with the warping-based \textit{X2Face} method \cite{X2Face} as well as the one-shot, image-based \textit{First Order Motion Model for Image Animation (FOMM)} \cite{Siarohin_2019_NeurIPS} system. As suggested visually in Fig. \ref{fig:fig11} and the quantitative results in Table \ref{table5}, \textit{Head2Head++} outperforms both few-shot methods by a large margin. Such a discrepancy can be explained by the fact that we train an individual person-specific model for each target identity, instead of relying on a few image samples of the subject. On the other hand, methods such as first order motion model capitalise on a single image to generate the target person, which results in the inherent ambiguity when the desired head pose is very different from the one displayed on that single reference image. Unavoidably, this makes the identity preservation problem  more prominent in one-shot models.

\begin{figure}[t!]
    \centering
    \includegraphics[width=3.5in]{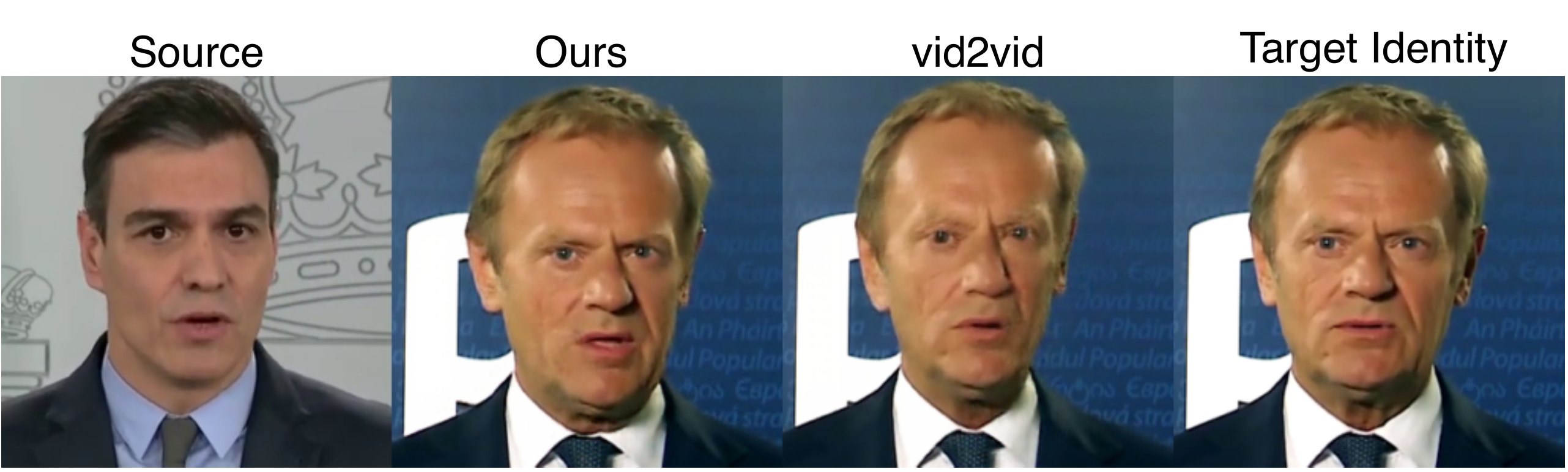}
    \caption{Comparison with \textit{vid2vid} conditioned on facial landmarks. Here, the identity preservation problem of landmarks is evident, as the head geometry of the source is reflected on the generated target.}
    \label{fig:fig12}
\end{figure}

In order to demonstrate the significance of the NMFC representation, we conducted a head reenactment experiment conditioning on landmarks. To that end, we trained \textit{vid2vid} \cite{vid2vid} on the task of landmark-to-RGB video translation. We demonstrate the results in Fig. \ref{fig:fig12}. As can be seen, the facial shape of the source has been transferred along with the pose and expression to the generated target.

\begin{figure}[t!]
\centering
\subfloat[]{\includegraphics[width=1.75in]{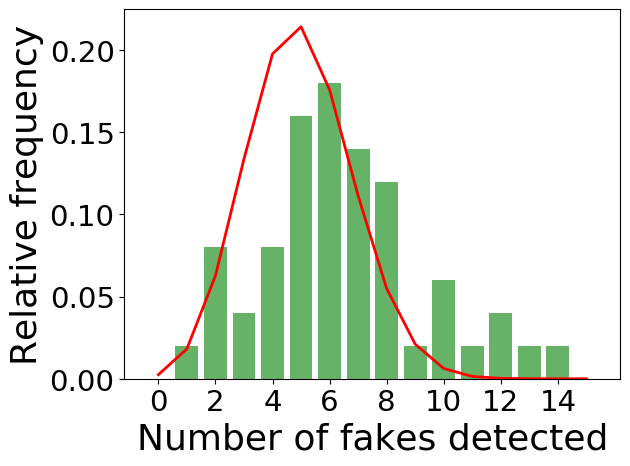}
\label{fig:fig14a}}
\subfloat[]{\includegraphics[width=1.75in]{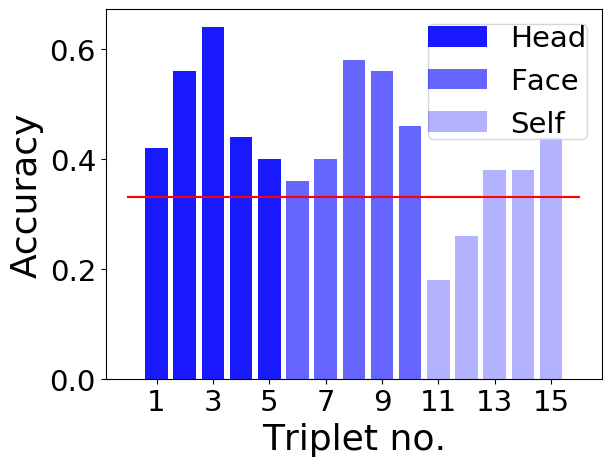}
\label{fig:fig14b}}
\caption{User study results. (a) The predictive performance of participants against random selection (Binomial distribution on $n=15$ independent Bernoulli trials with success probability $p=1/3$). (b) The fake detection accuracy of participants on each triplet of video samples.}
\label{fig:fig14}
\end{figure}

\subsection{Automated Study}
The progress recently made by generative deep learning methods is so impressive to the extent that manipulated videos (deepfakes) by such approaches are indistinguishable from the real ones. Consequently, a new research topic has emerged to tackle the detection of what is commonly known as `deepfakes'. A recent attempt in that direction is the \textit{Face Forensics++} \cite{roessler2019faceforensicspp}. The authors of  \cite{roessler2019faceforensicspp} collected a dataset of 1000 YouTube videos and manipulated them with graphics-based \cite{face2face, Marek2019} and learning-based \cite{torzdf2019, DeferredNeuralRendering}, facial reenactment methods. With this dataset, a CNN was trained to outrank the performance of human observers in detecting manipulated videos. Such a network was trained on $1.8$ million facially manipulated frames and reached a very high detection accuracy on the test split of their dataset (around $99 \%$). We use their pre-trained best-performing network and fine-tune it on around 14K frames synthesised from eight different subjects by our \textit{Head2Head++} approach and another 14K frames synthesised by \textit{Head2Head}. While fine-tunning, we set the learning-rate to 0.0001 and batch-size to 32. Adam optimiser \cite{adam} was also used with the default parameters. We then test the fine-tuned network on 3.2K frames generated by \textit{Head2Head++} and \textit{Head2Head} methods (1.6K frames each) and not seen during training. The trained network reports an accuracy of $76\%$ vs $80.2\%$  of fake frames detection for \textit{Head2Head++} and \textit{Head2Head}, respectively. This indicates that the fine-tuned forgery detection network finds it easier to spot fake frames generated by Head2Head compared to Head2Head++. Moreover, the same network performs way better (around $99 \%$) in detecting fake frames synthesised by any of the four head-reenactment methods tested in \cite{roessler2019faceforensicspp}.
Thanks to our carefully designed \textit{Head2Head++} framework, the trained forgery detection network finds it harder to identify our fake frames, which can be attributed to the increased realism of our results. 

\subsection{User Study}

We further evaluate the photo-realism of our generated videos, by conducting a user study. For that, we synthesised five videos of 75 frames (3 seconds) each, using different reenactment methods (head, face and self reenactment). Then, we coupled each fake video with two real ones, with the same duration and same target identity, forming triplets of videos. Next, we asked 50 participants on MTurk to detect the fake sample of each triplet. We presented the video triplet to the participants, allowing them to watch them only once.

Our recorded results indicate a human fake detection accuracy of $43.1 \%$, which demonstrates the strong photo-realism of samples created by our \textit{Head2Head++} system. In Fig. \ref{fig:fig14a}, we show the relative frequency of participants on the number of fake videos that were successfully detected. As can be  seen, the majority managed to spot between 4 and 8 out of the 15 synthetic samples we displayed. The comparison of our statistics with a Binomial distribution (red curve) indicates that user study participants performed slightly better than random selection. Nonetheless, a significant percentage of them, around $18\%$, managed to detect 9 or more fake samples. Finally, Fig. \ref{fig:fig14b} displays the predictive accuracy of participants on each video triplet. It is evident that self reenactment samples were indistinguishable from real ones, with just $32.8 \%$ predictive accuracy. Videos generated under face and head reenactment were identified a little more successfully, with an accuracy of $47.2\%$ and $49.2\%$ respectively.

\section{Conclusion and Future Work}

We proposed \textit{Head2Head++}, a pipeline consisting of a novel 3D facial reconstruction system and a Video Rendering Network, able to perform full head reenactment from a source to a target identity. Our new 3D face recovery and eye tracking methods allow our system to operate in nearly real-time speeds. The extensive set of experiments we performed demonstrated the generative abilities of our system, as well as the significance of its components. Furthermore, qualitative and quantitative comparisons suggest that \textit{Head2Head++} outperforms other state-of-the-art methods, in terms of photo-realism, expression and pose transferability, as well as target identity preservation. For future work, we aim to make the network trainable in a few-shot manner, reducing the training time and eliminating the need for learning a different model per person. Additionally, we plan to transfer the facial expressions from the source in a way that preserves the target's speaking style.



%



\ifCLASSOPTIONcaptionsoff
  \newpage
\fi



\bibliographystyle{IEEEtran}
\bibliography{IEEEabrv,bibliography.bib}

%



%

\begin{IEEEbiography}[{\includegraphics[width=1in,height=1.25in,clip,keepaspectratio]{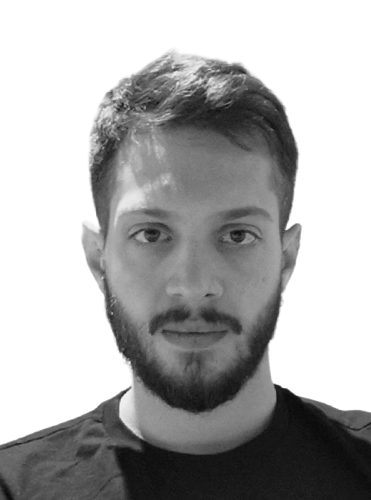}}]{Michail Christos Doukas}
is a PhD Student at Imperial College London and is currently doing his internship at Huawei UK Research Centre. He received the MSc degree in Computing from Imperial College London, in 2017. Prior to that, he has studied Electrical and Computer Engineering (MEng 2016) at the National Technical University of Athens (NTUA), Greece. His research interests are focused on Deep Learning and Computer Vision, including Generative Adversarial Neural Networks, Video-to-Video Translation, Visual Speech Synthesis, Face and Head Reenactment and Few-shot Learning.
\end{IEEEbiography}

\begin{IEEEbiography}[{\includegraphics[width=1in,height=1.25in,clip,keepaspectratio]{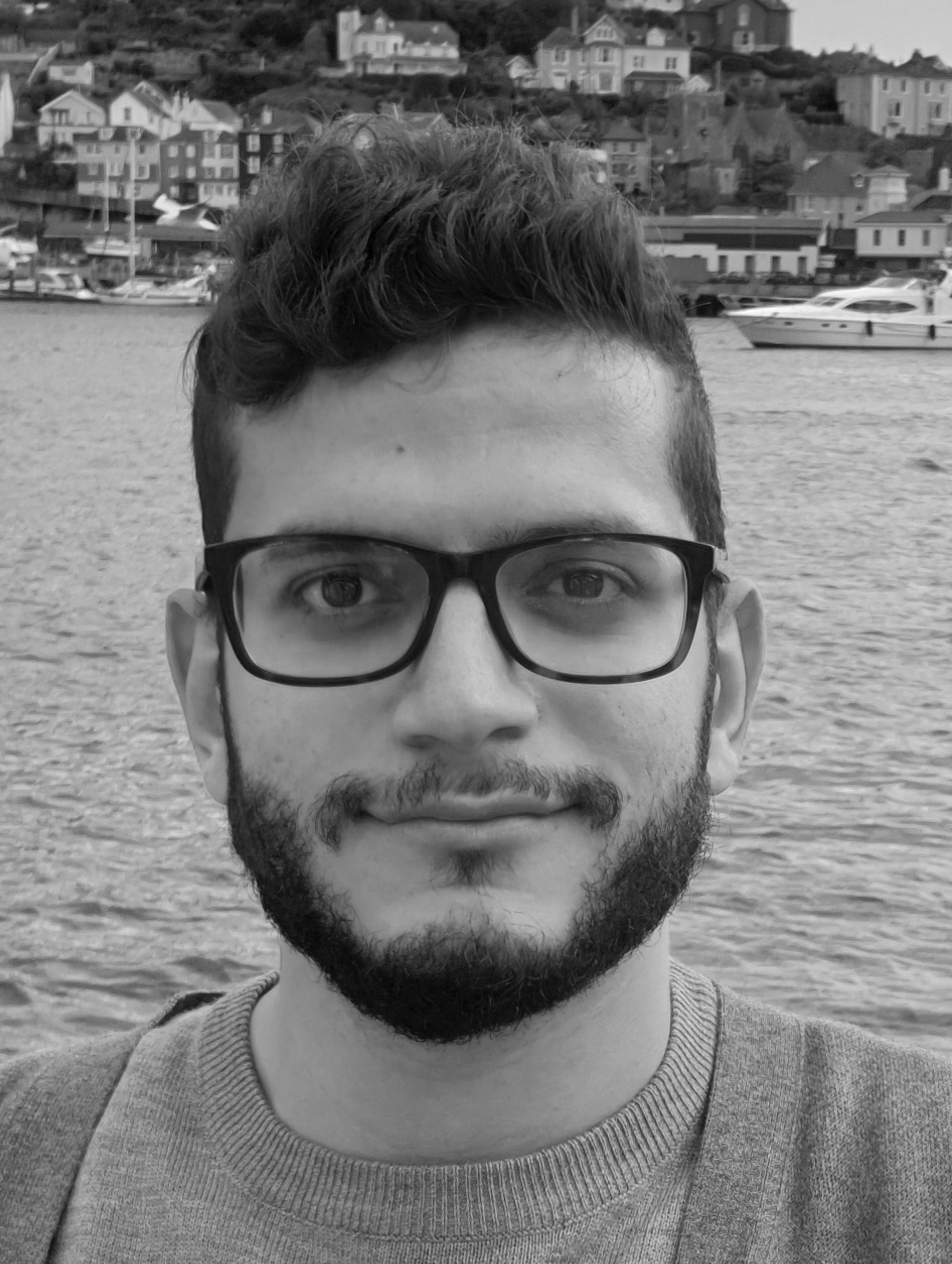}}]{Mohammad Rami Koujan}
is currently a PhD student at the University of Exeter (UK). Before his PhD, he did a 2-year Erasmus Mundus Joint MSc. course in Computer Vision and Robotics at Heriot-Watt University, University of Bourgogne and University of Girona and graduated with distinction. As an undergraduate student, he completed a Communications and Information engineering course with first class honors at Yarmouk Private University, Syria. His research interests lie mainly in 3D Computer Vision and Deep Learning techniques. He has been focusing on 3D facial shape modelling and analysis, non-rigid 3D facial motion capture and photo-realistic synthesis.
\end{IEEEbiography}

\begin{IEEEbiography}[{\includegraphics[width=1in,height=1.5in,clip,keepaspectratio]{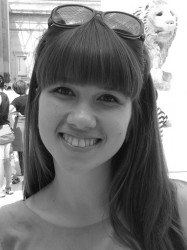}}]{Viktoriia Sharmanska}
is a research fellow at Imperial College London, leading the project in 'Deep Understanding of Human Behaviour from Video Data: Action plus Emotion Approach', since October 2017. Prior to her current position, she was a visiting research fellow at the University of Sussex, UK, working on cross-modal and cross-dataset learning with privileged information. She got her MSc in Applied Mathematics from the Taras Shevchenko National University of Kyiv, Ukraine, and her PhD in Computer Vision and Machine Learning from the Institute of Science and Technology Austria.  Her research interests include deep learning methods for understanding human behaviour from facial and bodily cues, algorithmic fairness, designing machine learning models that can overcome human and dataset collection biases. 
\end{IEEEbiography}


\begin{IEEEbiography}[{\includegraphics[width=1in,height=1.25in,clip,keepaspectratio]{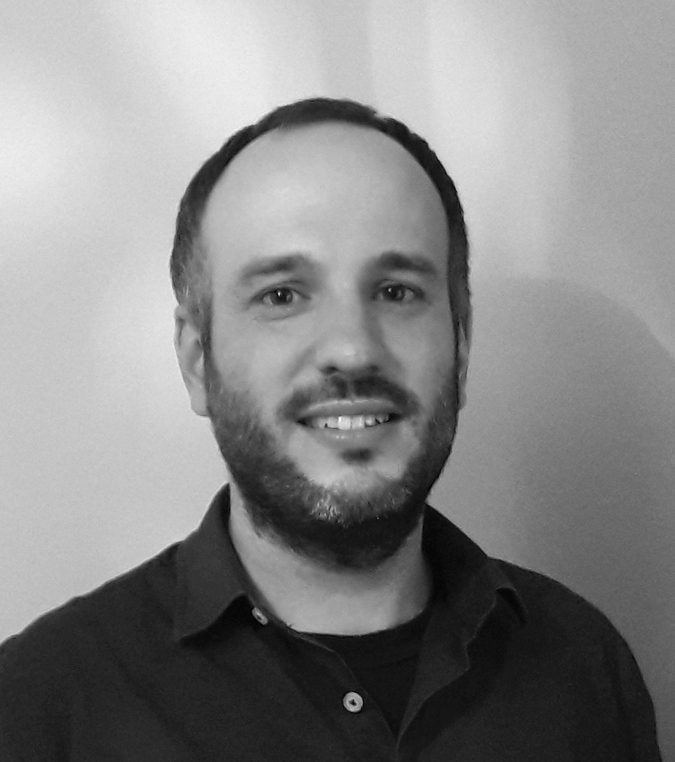}}]{Anastasios Roussos}
is a Principal Researcher (Associate Professor level) at the Foundation for Research and Technology - Hellas (FORTH), Greece. He is also an Honorary Senior Lecturer in the University of Exeter, UK. Prior to his current positions, he was a Lecturer in Computer Science at the University of Exeter and a Fellow of the Alan Turing Institute, UK. Before these positions, he was a postdoctoral researcher at Imperial College London, University College London and Queen Mary, University of London. He has studied Electrical and Computer Engineering (PhD 2010, Dipl-Ing 2005) at the National Technical University of Athens (NTUA), Greece. His research specialises in the fields of Computer Vision and Machine Learning. 
\end{IEEEbiography}

\begin{IEEEbiography}[{\includegraphics[width=1in,height=1.25in,clip,keepaspectratio]{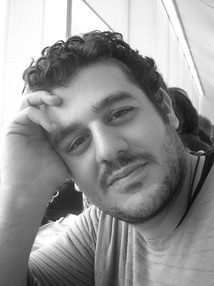}}]{Stefanos Zafeiriou}
is currently a Professor in Machine Learning and Computer Vision with the Department of Computing, Imperial College London, London, U.K, and an EPSRC Early Career Research Fellow. Between 2016-2020 he was also a Distinguishing Research Fellow with the University of Oulu under Finish Distinguishing Professor Programme. He was a recipient of the Prestigious Junior Research Fellowships from Imperial College London in 2011. He was the recipient of the President’s Medal for Excellence in Research Supervision for 2016. His research specialises in machine learning methodologies applied to computer vision problems, such as 2-D/3-D face analysis, deformable object fitting and tracking, shape from shading, and human behaviour analysis.
\end{IEEEbiography}




\end{document}